\documentclass[runningheads]{llncs}

\usepackage{eccv}

\usepackage{eccvabbrv}

\usepackage{graphicx}
\usepackage{booktabs}

\usepackage[accsupp]{axessibility}  % Improves PDF readability for those with disabilities.

\usepackage[pagebackref,breaklinks,colorlinks,citecolor=eccvblue]{hyperref}
\usepackage{orcidlink}

\usepackage{algorithm2e}
\RestyleAlgo{ruled}
\SetKwComment{Comment}{/* }{ */}
\newcommand{\plusequals}{\mathrel{+}=}
\newcommand{\timesequals}{\mathrel{*}=}

\newcommand{\veryshortarrow}[1][3pt]{\mathrel{%
   \hbox{\rule[\dimexpr\fontdimen22\textfont2-.2pt\relax]{#1}{.4pt}}%
   \mkern-4mu\hbox{\usefont{U}{lasy}{m}{n}\symbol{41}}}}
   
\usepackage[dvipsnames]{xcolor}
\usepackage{bbm}
\usepackage{dsfont}

\newcommand{\ShortName}{\textsc{NeRF-XL}\xspace}

\title{NeRF-XL: Scaling NeRFs with Multiple GPUs}

\author{Ruilong Li\inst{1,2} \and
Sanja Fidler\inst{1} \and
Angjoo Kanazawa\inst{2} \and
Francis Williams\inst{1}}

\authorrunning{R.~Li et al.}
\institute{~\inst{1}NVIDIA \quad\quad\quad\quad\quad \inst{2}UC Berkeley}

\begin{document}
% \maketitle

\maketitle
\vspace{-1mm}
\vspace{-4mm}
\begin{abstract}
We present \ShortName, a principled method for distributing Neural Radiance Fields (NeRFs) across multiple GPUs, thus enabling the training and rendering of NeRFs with an arbitrarily large capacity. We begin by revisiting existing multi-GPU approaches, which decompose large scenes into multiple independently trained NeRFs~\cite{tancik2022blocknerf,Turki_2022_CVPR,meuleman2023progressively}, and identify several fundamental issues with these methods that hinder improvements in reconstruction quality as additional computational resources (GPUs) are used in training. \ShortName remedies these issues and enables the training and rendering of NeRFs with an arbitrary number of parameters by simply using more hardware. At the core of our method lies a novel distributed training and rendering formulation, which is mathematically equivalent to the classic single-GPU case and minimizes communication between GPUs. By unlocking NeRFs with arbitrarily large parameter counts, our approach is the first to reveal multi-GPU scaling laws for NeRFs, showing improvements in reconstruction quality with larger parameter counts and speed improvements with more GPUs. We demonstrate the effectiveness of \ShortName on a wide variety of datasets, including the largest open-source dataset to date, MatrixCity~\cite{li2023matrixcity}, containing 258K images covering a 25km$^2$ city area. Visit our webpage at \url{https://research.nvidia.com/labs/toronto-ai/nerfxl/} for code and videos.
\end{abstract}
\vspace{-4mm}

\begin{figure*}[!h]
\vspace{-5mm}
\begin{center}
\includegraphics[width=0.98\linewidth]{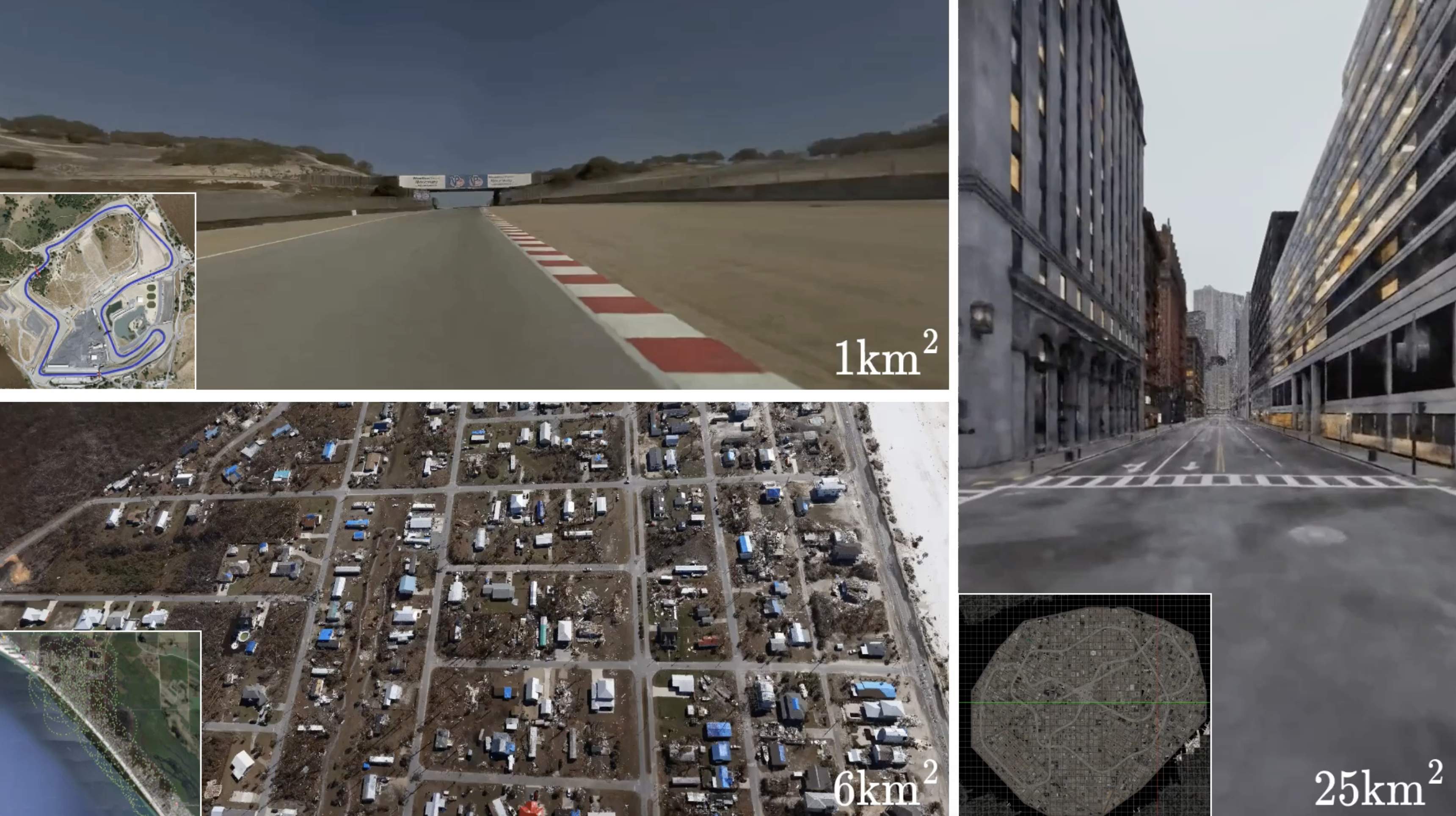}
\vspace{-1mm}
% \caption{Extremely large scale NeRFs reconstructed by our principled multi-GPU distributed algorithm.}
\caption{Our principled multi-GPU distributed training algorithm enables scaling up NeRFs to arbitrarily-large scale.}
\end{center}
\vspace{-5mm}
\end{figure*}

\section{Introduction}

%P1 (Motivation/Evokes the application)
\looseness=-1 Recent advances in novel view synthesis have greatly enhanced our ability to capture Neural Radiance Fields (NeRFs), making the process significantly more accessible. These advancements enable the reconstruction of both larger scenes and finer details within a scene. Expanding the scope of a captured scene, whether by increasing the spatial scale (e.g., capturing a multi-kilometer-long cityscape) or the level of detail (e.g., scanning the blades of grass in a field), involves incorporating a greater volume of information into the NeRF for accurate reconstruction. Consequently, for scenes with high information content, the number of trainable parameters required for reconstruction may exceed the memory capacity of a single GPU.

% P2: What we do
\looseness=-1 In this paper, we introduce \ShortName, a principled algorithm for efficiently distributing Neural Radiance Fields (NeRFs) across multiple GPUs. Our method enables the capture of high-information-content scenes, including those with large-scale and high-detail features, by simply adding more hardware resources. At its core, \ShortName allocates NeRF parameters across a disjoint set of spatial regions and trains them jointly across GPUs. Unlike conventional distributed training pipelines that synchronize gradients during the backward pass, our approach only requires information synchronization during the forward pass. Additionally, we drastically reduce the required data transfer between GPUs by carefully rewriting the volume rendering equation and relevant loss terms for the distributed setting. This novel rewriting enhances both training and rendering efficiency. The flexibility and scalability of our approach allows us to efficiently optimize NeRFs with an arbitrary number of parameters using multiple GPUs.

% P3: In contrast
\looseness=-1 Our work contrasts with recent approaches that utilize multi-GPU algorithms to model large-scale scenes by training a set of independent NeRFs~\cite{tancik2022blocknerf, Turki_2022_CVPR, meuleman2023progressively}. While these approaches require no communication between GPUs, each NeRF needs to model the entire space, including the background region. This leads to increased redundancy in the model's capacity as the number of GPUs grows. Additionally, these methods require blending NeRFs during rendering, which degrades visual quality and introduces artifacts in overlapping regions. Consequently, unlike \ShortName, these methods fail to achieve visual quality improvements as more model parameters (equivalent to more GPUs) are used in training.

We demonstrate the effectiveness of our method across a diverse set of captures, including street scans, drone flyovers, and object-centric videos. These range from small scenes (10m$^2$) to entire cities (25km$^2$). Our experiments show that \ShortName consistently achieves improved visual quality (measured by PSNR) and rendering speed as we allocate more computational resources to the optimization process. Thus, \ShortName enables the training of NeRFs with arbitrarily large capacity on scenes of any spatial scale and detail.
\section{Related Work}

\paragraph{Single GPU NeRFs for Large-Scale Scenes}

Many prior works have adapted NeRF to large-scale outdoor scenes. For example, BungeeNeRF~\cite{xiangli2022bungeenerf} uses a multi-scale, coarse-to-fine pipeline to address memory constraints; Grid-guided NeRF~\cite{xu2023grid} uses multiple image planes for drone-captured footage; F2-NeRF~\cite{wang2023f2} introduces a space warping algorithm for efficient level-of-detail handling in a free camera trajectory capture; and UrbanNeRF~\cite{rematas2022urban} leverages LiDAR and segmentation maps to improve in-the-wild captures. Despite their advancements, these prior works are bounded by the computational capacity of a single GPU.

\paragraph{NeRFs with Multiple GPUs}
An alternative approach for training NeRFs on large-scale scenes is to use multiple GPUs. 
BlockNeRF~\cite{tancik2022blocknerf}, MegaNeRF~\cite{turki2022mega} and SNISR~\cite{wu2022scalable} partition a scene into overlapping NeRFs based on camera trajectory or spatial content, and optimize each NeRF independently (one per GPU).
ProgressiveNeRF~\cite{meuleman2023progressively} adopts a similar strategy but recursively optimizes one NeRF at a time with overlapped blending.
While these methods overcome the memory limitations of a single GPU, each independent NeRF has to model the entire scene within a spatial region, leading to increased redudancy (in the model's capacity) and decreased visual quality as more GPUs are used in training. Furthermore, these methods must rely on depth initialization for spatial partitioning~\cite{wu2022scalable}, or introduce overlapping between NeRFs~\cite{tancik2022blocknerf, turki2022mega, meuleman2023progressively}, which causes visual artifacts during rendering. We provide an in-depth analysis of the problems faced by  prior multi-GPU methods in the next section.
\section{Revisiting Existing Approaches: Independent Training}
\label{sec:revist}

\begin{figure*}[!t]
\centering
\includegraphics[width=1.0\linewidth]{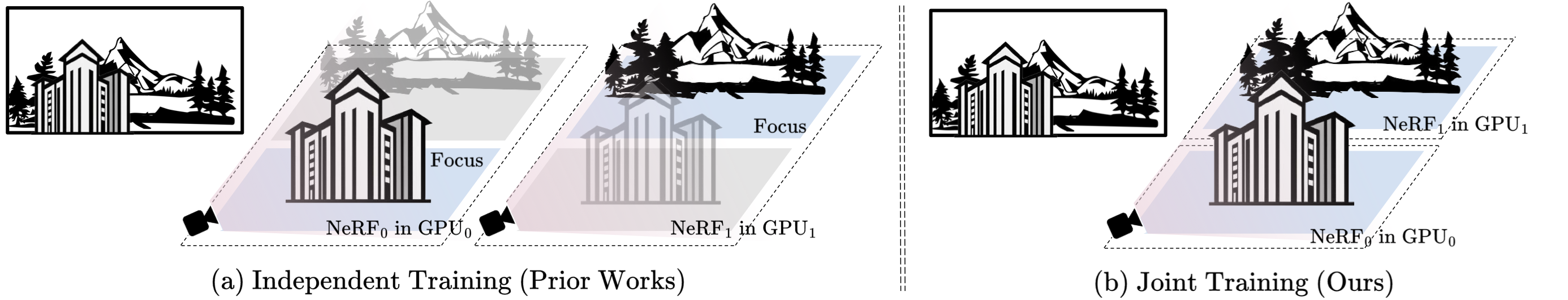}
\vspace{-2mm}
\caption{
\textbf{Independent Training v.s. Joint Training with multi-GPU.} 
Training multiple NeRFs independently~\cite{tancik2022blocknerf, turki2022mega, meuleman2023progressively} requires each NeRF to model both the focused region and its surroundings, leading to redundancy in model's capacity. In contrast, our joint training approach utilizes non-overlapping NeRFs, thus without any redundancy.
}
\vspace{-3mm}
\label{fig:prior_work_illustration}
\end{figure*}

\begin{figure}[!t]
\centering
\includegraphics[width=0.90\linewidth]{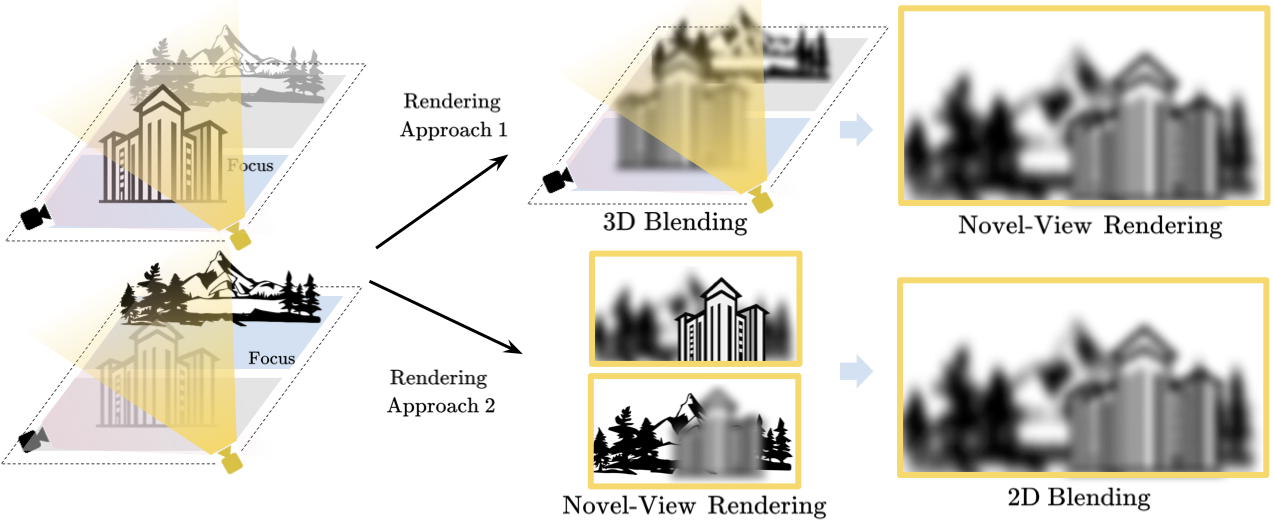}
\caption{
\textbf{Independent Training requires Blending for Novel-View Synthesis.} Either blending in 2D~\cite{tancik2022blocknerf,meuleman2023progressively} or 3D~\cite{turki2022mega} introduces blurriness into the rendering.
}
\vspace{-2mm}
\label{fig:rendering_issue_illustration}
\end{figure}

\begin{figure}[!t]
\centering
\includegraphics[width=0.95\linewidth]{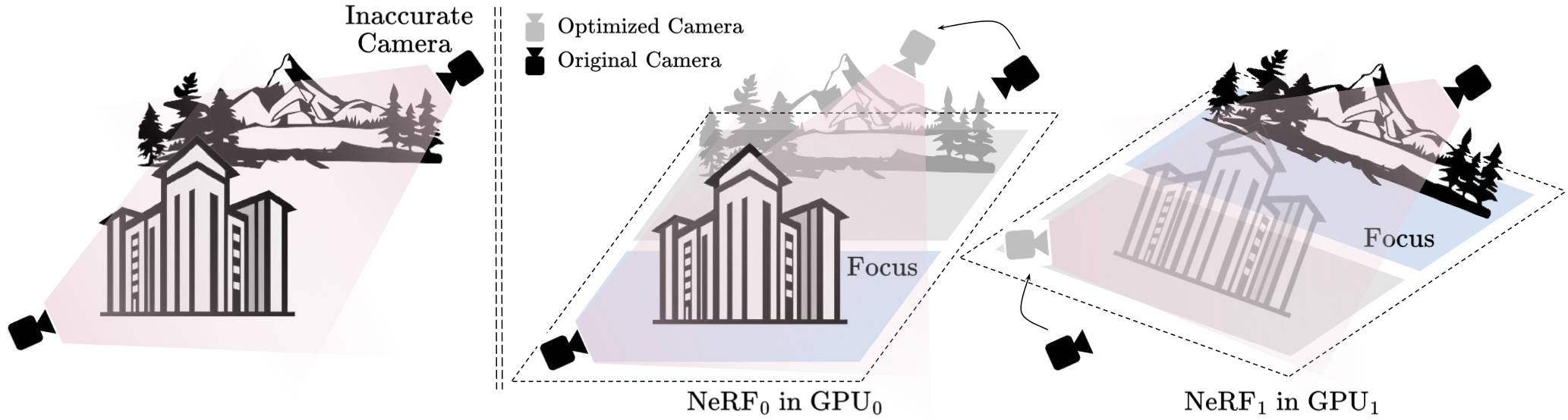}
\caption{
\textbf{Independent Training Creates Distinct Camera Optimizations.} Camera optimization in NeRF can be achieved by either transforming the inaccurate camera itself or all other cameras along with the underlying 3D scene. Thus, training multiple NeRFs independently with camera optimization may lead to inconsistencies in camera corrections and scene geometry, causing more difficulties for blended rendering.}
\vspace{-2mm}
\label{fig:camera_optimization_issue_illustration}
\end{figure}

In leveraging multiple GPUs for large-scale captures, prior research~\cite{meng2021gnerf, tancik2022blocknerf, meuleman2023progressively} has consistently employed the approach of training multiple NeRFs focusing on different spatial regions, where each NeRF is trained independently on its own GPU. 
\emph{However, independently training multiple NeRFs has fundamental issues that impede visual-quality improvements with the introduction of additional resources (GPUs).} 
%contradicts one of the core motivations of multi-GPU training for large-scale captures. 
%
This problem is caused by three main issues described below.

\paragraph{Model Capacity Redundancy.} 
The objective of training multiple independent NeRFs is to allow each NeRF to focus on a different (local) region and achieve better quality within that region than a single global model with the same capacity. Despite this intention, each NeRF is compelled to model not only its designated region but also the surrounding areas, since training rays often extend beyond the designated region as depicted in Figure~\ref{fig:prior_work_illustration}(a). This leads to an inherent redundancy in the model's capacity since each NeRF must model both the local and surrounding regions. As a result, increasing the number of GPUs (and hence using smaller spatial regions per NeRF), increases the total redundancy in the model's capacity. For example, Mega-NeRF~\cite{turki2022mega} exhibits 38\%/56\%/62\% ray samples outside the tiled regions with 2$\times$/4$\times$/8$\times$ tiles on the \textsc{University4} capture.
In contrast, our proposed method of jointly training all tiles removes the need for surrounding region modeling in each NeRF, \emph{thereby completely eliminating redundancy}, as shown in Figure~\ref{fig:prior_work_illustration}(b)). This feature is crucial for efficiently leveraging additional computational resources.

\paragraph{Blending for Rendering.} 

\begin{figure}[!t]
\centering
\includegraphics[width=1.0\linewidth]{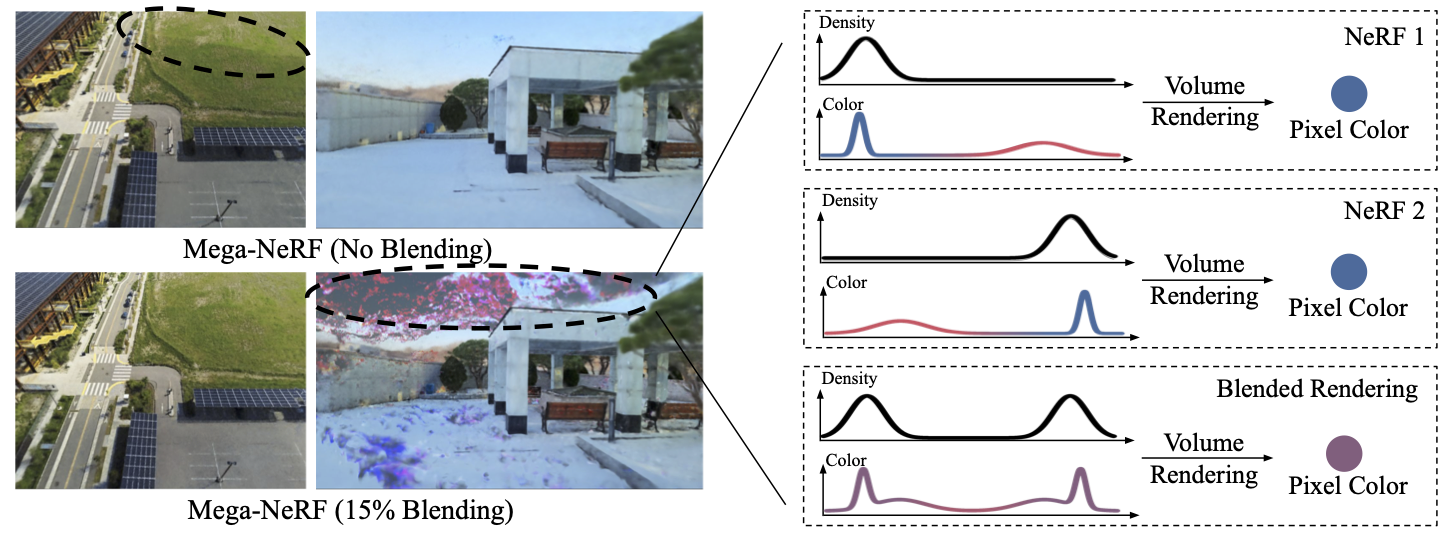}
\caption{\textbf{Potential Artifacts Caused by 3D Blending.} On the left we show Mega-NeRF results trained with 2 GPUs. At $0\%$ overlap, boundary artifacts appear due to independent training, while at $15\%$ overlap, severe artifacts appear due to 3D blending. On the right we illustrate the reason behind this artifact: while each independently trained NeRF renders the correct color, the blended NeRF do not guarantee correct color rendering.}
\vspace{-1mm}
\label{fig:3d_blending_artifacts}
\end{figure}

When rendering independently trained NeRFs, it is often necessary to employ a blending strategy to merge the NeRFs and mitigate inconsistencies at the region boundaries. Past works typically choose local regions with  a certain degree of overlap, such as 50\% in Block-NeRF~\cite{tancik2022blocknerf} and 15\% in Mega-NeRF~\cite{turki2022mega}. Two primary approaches exist for blending NeRFs during novel-view synthesis. One approach involves rendering each NeRF independently and then blending the resulting images when the camera is positioned within the overlapped region (referred to as 2D blending)~\cite{tancik2022blocknerf, meuleman2023progressively}. The alternative approach is to blend the color and density in 3D for ray samples within the overlapped region (referred to as 3D blending)~\cite{turki2022mega}. 
As illustrated in Figure~\ref{fig:rendering_issue_illustration}, both approaches can introduce blur into the final rendering. Moreover, blending in 3D can lead to more pronounced artifacts in rendering, due to deviations in the volume rendering equation, as demonstrated in Figure~\ref{fig:3d_blending_artifacts}. In contrast, our joint training approach does not rely on any blending for rendering. In fact, our method renders the scene in the exact same way during training and inference, thereby eliminating the train-test discrepancies introduced by past methods.

\paragraph{Inconsistent Per-camera Embedding.}

In many cases, we need to account for things like white balance, auto-exposure, or inaccurate camera poses in a capture. A common approach to model these factors is by optimizing an embedding for each camera during the training process, often referred to as appearance embedding or pose embedding~\cite{tancik2023nerfstudio, martin2021nerf, lin2021barf}. However, when training multiple NeRFs independently, each on its own GPU, the optimization process leads to independent refinements of these embeddings. This can result in inconsistent camera embeddings due to the inherently ambiguous nature of the task, as demonstrated in Figure~\ref{fig:camera_optimization_issue_illustration}.
Inconsistencies in appearance embeddings across NeRFs can result in disparate underlying scene colors, while inconsistencies in camera pose embeddings can lead to variations in camera corrections and the transformation of scene geometry. These disparities introduce further difficulties when merging the tiles from multiple GPUs for rendering. Conversely, our joint training approach allows optimizing a single set of per-camera embeddings (through multi-GPU synchronization), thus completely eliminating these issues. \\

\noindent Due to the issues listed above, prior works~\cite{tancik2022blocknerf, turki2022mega} which train multiple independent NeRFs do not effectively harness the benefits of additional computational resources (GPUs) as they scale up, as evidenced in our experiments (\S~\ref{sec:experiments}). As a result, we advocate for a novel \emph{joint} training approach. Without any heuristics, our approach gracefully enhances both visual quality and rendering speed as more GPUs are used in training, Moreover, our method reveals the multi-GPU scaling laws of NeRF for the first time. 
%in the multi-GPU setting on all kinds of data at any scale.

\section{Our Method: Joint Training}
\label{sec:our_method}

\begin{figure*}[!t]
\centering
\includegraphics[width=1.0\linewidth]{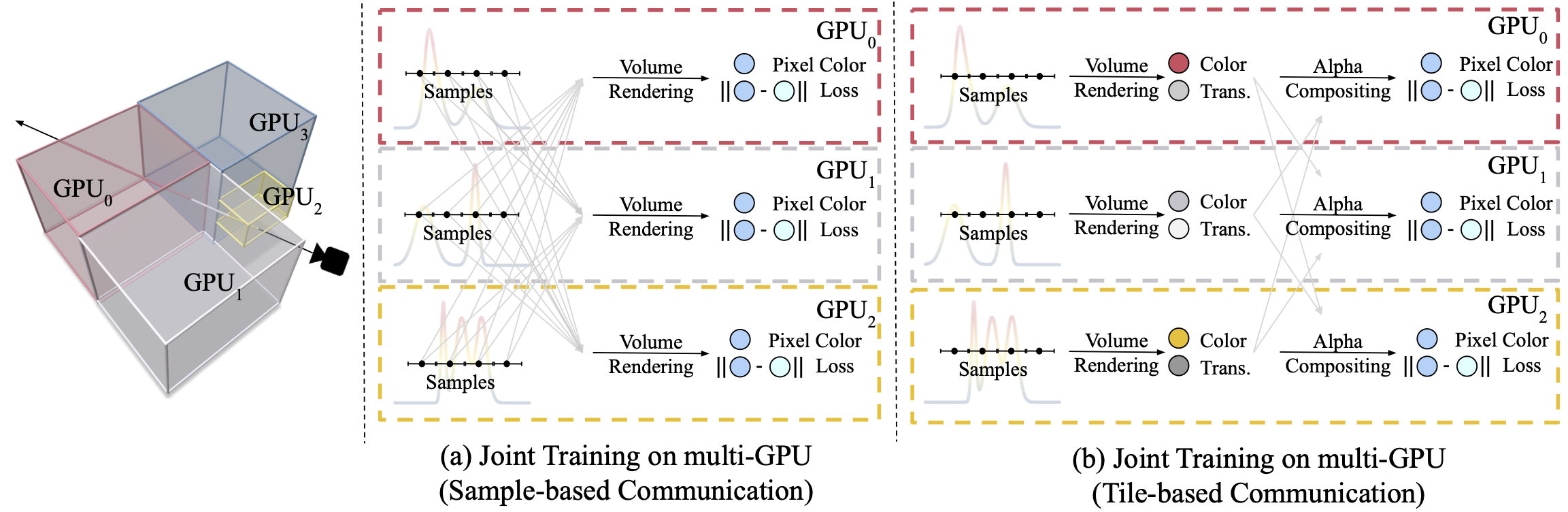}
\caption{\textbf{Our Training Pipeline.} Our method jointly trains multiple NeRFs across all GPUs, each of which covers a disjoint spatial region. The communication across GPUs only happens in the forward pass but not the backward pass (shown in gray arrows). (a) We can train this system by evaluating each NeRF to get the sample color and density, then broadcast these values to all other GPUs for a global volume rendering (\S~\ref{sec:method_naive}). (b) By rewriting volume rendering equation we can dramatically reduce the data transfer to one value per-ray, thus improving efficiency (\S~\ref{sec:method_efficient}).
}
\vspace{-1mm}
\label{fig:overview}
\end{figure*}

\subsection{Background}

\paragraph{Volume Rendering}
NeRFs~\cite{mildenhall2020nerf} employ volume rendering to determine the pixel color through the integral equation: 
\begin{equation}
\label{equ:volrend}
\begin{gathered}
C(t_n\veryshortarrow t_f) = \int_{t_n}^{t_f} T(t_n\veryshortarrow t) \sigma(t) c(t) \, dt,
\\
\quad \text{where} \quad
T(t_n\veryshortarrow t) = \exp\left(-\int_{t_n}^{t} \sigma(s) \, ds\right).
\end{gathered}
\end{equation}
Here, $T(t_n\veryshortarrow t)$ represents transmittance, $\sigma(t)$ denotes density, and $c(t)$ signifies the color at position $t$ along the ray.

\paragraph{Distortion Loss}
Initially introduced in Mip-NeRF 360~\cite{barron2022mip360} and validated in subsequent works~\cite{barron2023zipnerf,tancik2023nerfstudio,li2023nerfacc}, this loss acts as a regularizer to address ``floater'' artifacts in NeRF reconstructions caused by limited training viewpoint coverage. It is calculated along a ray as
\begin{equation}
\begin{aligned}
\mathcal{L}_{dist}(t_n\veryshortarrow t_f)
& = \int_{t_n}^{t_f} w(t_i) w(t_j) \left| t_i - t_j \right| \, dt_i \, dt_j ,
\end{aligned}
\end{equation}
where $w(t)=T(t_n\veryshortarrow t)\sigma(t)$ represents the volume rendering weight for each sample along the ray. Intuitively, it penalizes floaters by encouraging density concentration in minimal, compact regions. See~\cite{barron2022mip360} for more details. \\

\subsection{Non-overlapped NeRFs}
\label{sec:method_naive}
A straightforward strategy to increase model capacity with multiple GPUs is to partition 3D space into tiles and allocate a NeRF for each tile. But unlike prior works~\cite{turki2022mega, tancik2022blocknerf, meuleman2023progressively} that employ overlapped NeRFs to model both tiles and their surrounding regions, our method exclusively models \emph{non-overlapped} tiles, with each NeRF assigned to a single tile. This distinction is illustrated in Figure~\ref{fig:prior_work_illustration}.

To render our NeRFs across multiple GPUs, we first distribute ray samples among GPUs based on the bounding box of the tiles. Notably it's important to ensure that sample intervals do not extend beyond tile boundaries to prevent overlap between samples.
We subsequently query sample attributes (\ie color and density) on each respective GPU. Volume rendering is then performed through a global gather operation, consolidating information across all GPUs onto a single GPU to compute the final pixel color. Since all sample intervals are non-overlapping, the scene can be rendered accurately following the volume rendering equation without the need for any blending.

Training proceeds in a similar fashion to rendering, except that during the forward pass \emph{each} GPU performs the global gather operation (\ie broadcast) to obtain the information (\ie color and density) from all other GPUs (illustrated as gray lines in Figure~\ref{fig:overview}(a)). Then, each GPU computes the loss locally and back-propagates the gradients to its own parameters. Notably the forward pass produces the exact same loss values on every GPU, but each loss lives in a different computational graph that only differentiates with respect to its own local parameters, thus no gradient communication are required across GPUs.

Such a naive scheme is extremely simple to implement, and mathematically identical to training and rendering a NeRF represented by multiple small NeRFs~\cite{reiser2021kilonerf, rebain2021derf} on a single large GPU. Distributing learnable parameters and computational cost across multiple GPUs allows scaling NeRF to scenes of any size, as well as making most parts of training fully parallel (\textit{e.g.}, network evaluation, back-propagation). Despite its simplicity and scalability in comparison to blending overlapping NeRFs in prior works~\cite{tancik2022blocknerf, turki2022mega, meuleman2023progressively}, this naive approach requires synchronizing $\mathcal{O}(SK^2)$ data across GPUs, where $K$ is the number of GPUs, and $S$ is the average number of samples per-ray per-GPU. As the number of GPUs increases or the ray step size decreases, synchronization across GPUs quickly becomes a bottleneck. Therefore, on top of this approach, we present a sophisticated solution that significantly alleviates the burden of multi-GPU synchronization in a principled manner.

\subsection{Partition Based Volume Rendering}
\label{sec:method_efficient}

If we consider the near-far region $[t_n\veryshortarrow t_f]$ consisting of $N$ segments $[t_1\veryshortarrow t_2, t_2\veryshortarrow t_3, ..., t_N\veryshortarrow t_{N+1}]$, we can rewrite the volume-rendering integral~\eqref{equ:volrend} into a sum of integrals for each segment along the ray:
\begin{equation}
\label{equ:accum_color}
\begin{aligned}
C(t_1\veryshortarrow t_{N+1})
& = \int_{t_1}^{t_{N+1}} T(t_1\veryshortarrow t)\sigma(t)c(t)dt = \sum_{k=1}^{N}T(t_1\veryshortarrow t_k)C(t_k\veryshortarrow t_{k+1}) \\
\end{aligned}
\end{equation}
in which the transmittance $T(t_1\veryshortarrow t_k)$ can be written as:
\begin{equation}
\label{equ:prefix_trans}
T(t_1\veryshortarrow t_{k}) = \prod_{i=1}^{k-1} T(t_i\veryshortarrow t_{i+1})
\end{equation}
The above equation states that volume rendering along an entire ray is equivalent to first performing volume rendering independently within each segment,
then performing alpha compositing on all the segments. 
We can also rewrite the accumulated weights $A(t_1\veryshortarrow t_{N+1})$ and  depths $D(t_1\veryshortarrow t_{N+1})$ in a similar way:
\begin{equation}
\label{equ:accum_weight}
\begin{aligned}
A(t_1\veryshortarrow t_{N+1})
& = \int_{t_1}^{t_{N+1}} T(t_1\veryshortarrow t)\sigma(t)dt = \sum_{k=1}^{N}T(t_1\veryshortarrow t_k)A(t_k\veryshortarrow t_{k+1}) \\
\end{aligned}
\end{equation}

\begin{equation}
\label{equ:accum_depth}
\begin{aligned}
D(t_1\veryshortarrow t_{N+1})
& = \int_{t_1}^{t_{N+1}} T(t_1\veryshortarrow t)\sigma(t)tdt = \sum_{k=1}^{N}T(t_1\veryshortarrow t_k)D(t_k\veryshortarrow t_{k+1}) \\
\end{aligned}
\end{equation}

We can further rewrite the point-based integral in the distortion loss as an accumulation across segments:
\begin{equation}
\label{equ:accum_dist}
\begin{aligned}
\mathcal{L}_{dist}(t_1\veryshortarrow t_{N+1})
& = \int_{t_1}^{t_{N+1}} w(t_i) w(t_j) \left| t_i - t_j \right| dt_i dt_j \\
& = 2\sum_{k=1}^{N}T(t_1\veryshortarrow t_k)S(t_1\veryshortarrow t_k) + \sum_{k=1}^{N}T(t_1\veryshortarrow t_k)^2 \mathcal{L}_{dist}(t_k\veryshortarrow t_{k+1})
\end{aligned}
\end{equation}
in which the $S(t_1\veryshortarrow t_k)$ is defined as:
\begin{equation}
\label{equ:s}
\begin{aligned}
S(t_1\veryshortarrow t_k)
& = D(t_k\veryshortarrow t_{k+1})A(t_1\veryshortarrow t_k)  - A(t_k\veryshortarrow t_{k+1})D(t_1\veryshortarrow t_k)
\end{aligned}
\end{equation}
Intuitively, the first term $S(t_1\veryshortarrow t_k)$ penalizes multiple peaks across segments (zero if only one segment has non-zero values), while the second term $\mathcal{L}_{dist}(t_k\veryshortarrow t_{k+1})$ penalizes multiple peaks within the same segment. This transforms the pairwise loss on all samples into a hierarchy: pairwise losses within each segment, followed by a pairwise loss on all segments. Derivations for all the above formulae are given in the appendix.

Recall that the main drawback of our naive approach was an expensive per-sample data exchange across all GPUs. The above formulae convert sample-based composition to tile-based composition. This allows us to first reduce the per-sample data into per-tile data in parallel within each GPU and exchange only the per-tile data across all GPUs for alpha compositing. This operation is cost-effective, as now the data exchange is reduced from $O(KS^2)$ to $O(S^2)$ (each GPU contains a single tile). Figure~\ref{fig:overview}(b) shows an overview of our approach. \S~\ref{sec:abl} quantifies the improvement gained from this advanced approach compared to the naive version.

In addition to the volume rendering equation and distortion loss, a wide range of loss functions commonly used in NeRF literature can be similarly rewritten to suit our multi-GPU approach. For further details, we encourage readers to refer to the appendix.

\subsection{Spatial Partitioning}
\label{sec:spatial_partitioning}
Our multi-GPU strategy relies on spatial partitioning, raising the question of how to create these tiles effectively. 
Prior works~\cite{tancik2022blocknerf,turki2022mega} opt for a straightforward division of space into uniform-sized blocks within a global bounding box. While suitable for near-rectangular regions, this method proves suboptimal for free camera trajectories and can lead to unbalanced compute assignment across GPUs. As noted in~\cite{wang2023f2}, a free camera trajectory involves uneven training view coverage, resulting in varying capacity needs across space (\textit{e.g.}, the regions that are far away from any camera require less capacity than regions near a camera). To achieve balanced workload among GPUs, we want to ensure each GPU runs a similar number of network evaluations (\ie has a similar number of ray samples). This balance not only allocates compute resources evenly but also minimizes waiting time during multi-GPU synchronization for communicating the data, as unequal distribution can lead to suboptimal GPU utilization. 

We propose an efficient partitioning scheme aimed at evenly distributing workload across GPUs. When a sparse point cloud is accessible (\textit{e.g.}, obtained from SFM), we partition the space based on the point cloud to ensure that each tile contains a comparable number of points. This is achieved by recursively identifying the plane where the Cumulative Distribution Function (CDF) equals 0.5 for the 3D point distribution along each axis. As a result, this approach leads to approximately evenly distributed scene content across GPUs.
In cases where a sparse point cloud is unavailable, indicating a lack of prior knowledge about the scene structure, we instead discretize randomly sampled training rays into 3D samples. This serves as an estimation of the scene content distribution based on the camera trajectory, enabling us to proceed with partitioning in a similar manner. This process is universally applicable to various types of captures, including street, aerial, and object-centric data, and runs very quickly in practice (typically within seconds). Please refer to the appendix for visualizations of partitioned tiles on different captures.

\section{Experiments}
\label{sec:experiments}

\begin{table}[t]
\centering
\setlength\tabcolsep{3pt} % default value: 6pt
\scalebox{0.85}{
\begin{tabular}{lcccccc}
\toprule
                    & Garden~\cite{barron2022mip360} & University4~\cite{meuleman2023progressively} & Building~\cite{turki2022mega} & Mexico Beach~\cite{civilair} & Laguna Seca & MatrixCity~\cite{li2023matrixcity} \\
\midrule
\#Img          & 161    & 939     & 1940   & 2258   & 27695   & 258003      \\
\#Pix$_{c}$    & 175M     & 1947M     & 1920M   & 2840M   & 47294M & 25800M      \\
\#Pix$_{d}$    & 0.84M     & 3.98M     & -      & 9.63M   & 2819M  & 2007M      \\
\bottomrule
\end{tabular}
}
\vspace{1mm}
\caption{
\textbf{Data Statistics.} Our experiments are conducted on these captures from various sources, including street captures (\textsc{University4}, \textsc{MatrixCity}, \textsc{Laguna Seca}), aerial captures (\textsc{Building}, \textsc{Mexico Beach}) and an object-centric 360-degree capture (\textsc{Garden}). These data span a wide range of scales, enabling a comprehensive evaluation of the multi-GPU system. Pix$_{c}$ and Pix$_{d}$ are denoted for color pixels and depth pixels, respectively.
}
\vspace{-4mm}
\label{tab:data_stats}
\end{table}

\paragraph{Datasets.} 
The crux of a multi-GPU strategy lies in its ability to consistently improve performance across all types of captures, regardless of scale, as additional resources are allocated. However, prior works typically evaluate their methods using only a single type of capture (\textit{e.g.}, street captures in Block-NeRF, aerial captures in Mega-NeRF). In contrast, our experiments are conducted on diverse captures from various sources, including street captures (\textsc{University4}~\cite{meuleman2023progressively}, \textsc{MatrixCity}~\cite{li2023matrixcity}, \textsc{Laguna Seca}\footnote{\label{laguna}Laguna Seca: An in-house capture of a 3.6km race track.}), aerial captures (\textsc{Building}~\cite{turki2022mega}, \textsc{Mexico Beach}~\cite{civilair}) and an object-centric 360-degree capture (\textsc{Garden}~\cite{barron2022mip360}). These data also span a wide range of scales, from \textsc{Garden} with 161 images in a 10m$^2$ area, to \textsc{MatrixCity} with 258K images in a 25km$^2$ area, thereby offering a comprehensive evaluation of the multi-GPU system. Table~\ref{tab:data_stats} shows detailed statistics for each of these captures.

\subsection{Joint Training v.s. Independent Training}

\begin{figure*}[!t]
\centering
\includegraphics[width=1.00\linewidth]{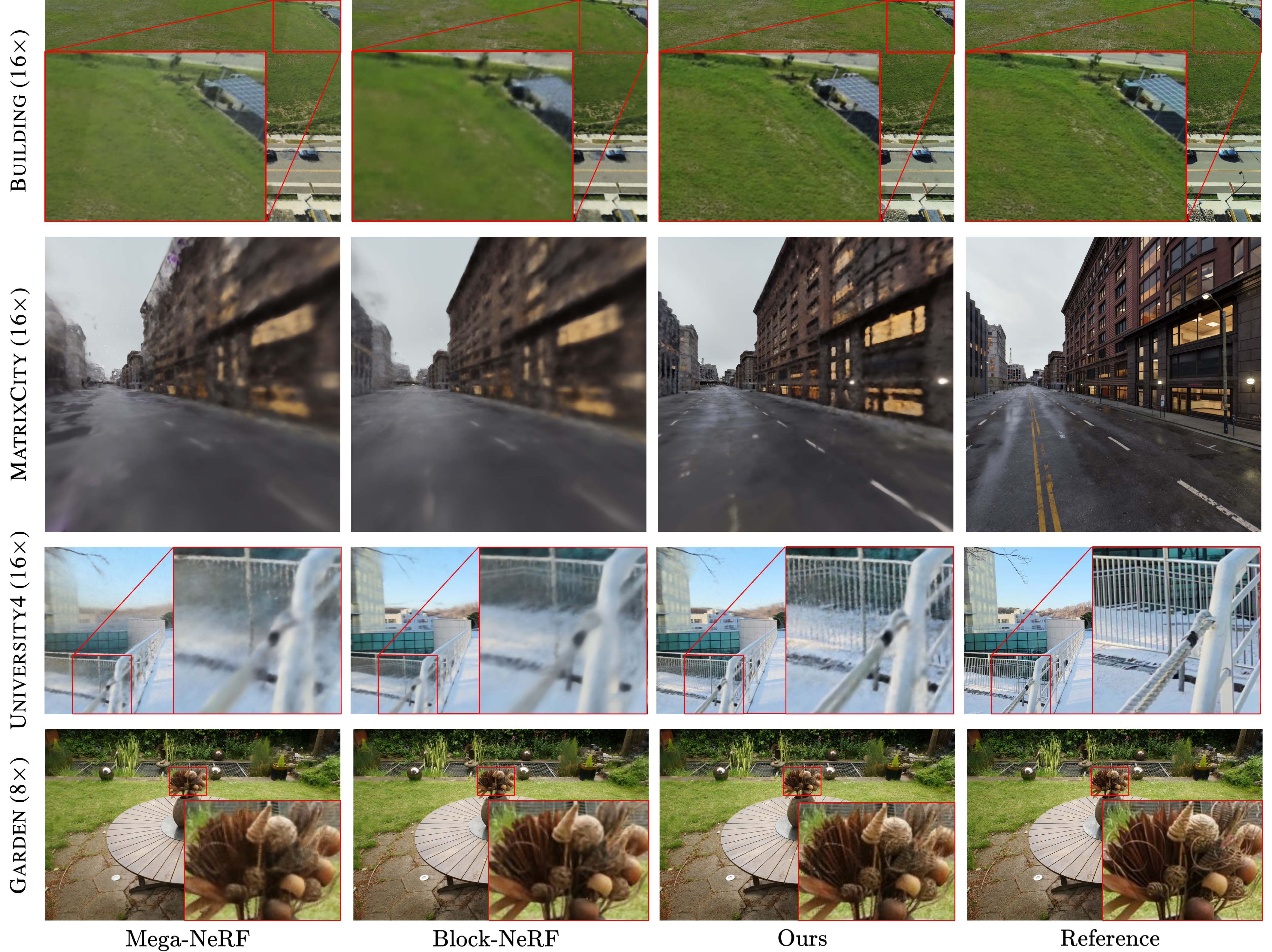}
\caption{\textbf{Qualitative Comparison.} Comparing to prior works, our method efficiently harnesses multi-GPU setups for performance improvement on all types of data.
} 
\label{fig:visual_comparision}
\end{figure*}

\begin{figure*}[!t]
\centering
\includegraphics[width=1.00\linewidth]{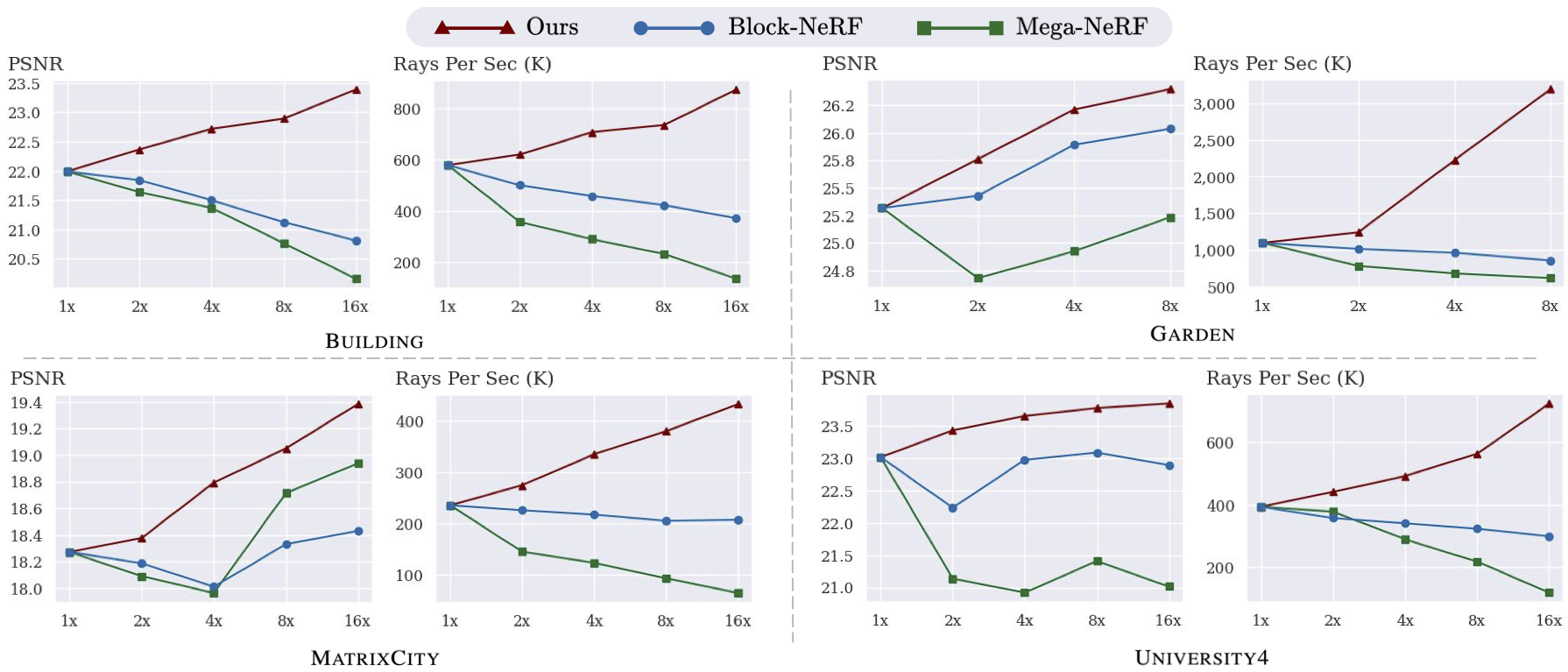}
\caption{\textbf{Quantitative Comparison.} Prior works based on independent training fails to realize performance improvements with additional GPUs, while our method enjoys improved rendering quality and speed as more resources are added to training.}
\label{fig:curves}
\end{figure*}

In this section, we conduct a comparative analysis between our proposed approach and two prior works, Block-NeRF~\cite{tancik2022blocknerf} and Mega-NeRF~\cite{turki2022mega}, all of which are aimed at scaling up NeRFs beyond the constraints of a single GPU. To ensure a fair evaluation solely on the multi-GPU strategies, we re-implemented each baseline alongside our method within a unified framework\footnote{On \textsc{Building} scene, our 8 GPU Mega-NeRF implementation achieves 20.8 PSNR comparing to 20.9 PSNR reported in the original paper.}. Each method is configured with the same NeRF representation (Instant-NGP~\cite{mueller2022instant}), spatial skipping acceleration structure (Occupancy Grid~\cite{mueller2022instant}), distortion loss~\cite{barron2022mip360}, and multi-GPU parallel inference. This standardized setup allows us to focus on assessing the performance of different multi-GPU strategies in both training (\ie, joint vs. independent~\cite{tancik2022blocknerf, turki2022mega}) and rendering (\ie, joint vs. 2D blending~\cite{tancik2022blocknerf} vs. 3D blending~\cite{turki2022mega}). For each baseline method, we adopt their default overlapping configurations, which is 15\% for Mega-NeRF and 50\% for Block-NeRF. All methods are trained for the same number of iterations ($20$K), with an equal number of total samples per iteration (effectively the batch size of the model).
Please refer to the appendix for implementation details.

In this section we conduct experiments on four captures, including \textsc{Garden}~\cite{barron2022mip360}, \textsc{Building}~\cite{turki2022mega}, \textsc{University4}~\cite{meuleman2023progressively} and \textsc{MatrixCity}~\cite{li2023matrixcity}, with GPU configurations ranging from 1$\times$ to 16$\times$ (multi-node). We evaluate the scalability of each method using two key metrics: Peak Signal-to-Noise Ratio (PSNR) for quality assessment and Rays Per Second for rendering speed, on the respective test sets of each capture.
As illustrated in Figure~\ref{fig:curves}, baseline approaches struggle to improve rendering quality with an increase in the number of GPUs, largely due to the inherent issues associated with independent training outlined in \S~\ref{sec:revist}. Additionally, baseline methods also fails to achieve faster rendering with additional GPUs, as they either need to evaluate duplicate pixels for 2D blending \cite{tancik2022blocknerf} or duplicate 3D samples for 3D blending \cite{turki2022mega}.
In contrast, our proposed approach, employing joint training and rendering, effectively eliminates model redundancy and train-test discrepancy. Thus, it gracefully benefits from increased parameters and parallelization with additional GPUs, resulting in nearly linear improvements in both quality and rendering speed. More qualitative comparisons are shown in Figure~\ref{fig:visual_comparision}.

\subsection{Robustness and Scalability}
\label{sec:robustness_and_scalability}

\begin{figure*}[!t]
\centering
\includegraphics[width=1.00\linewidth]{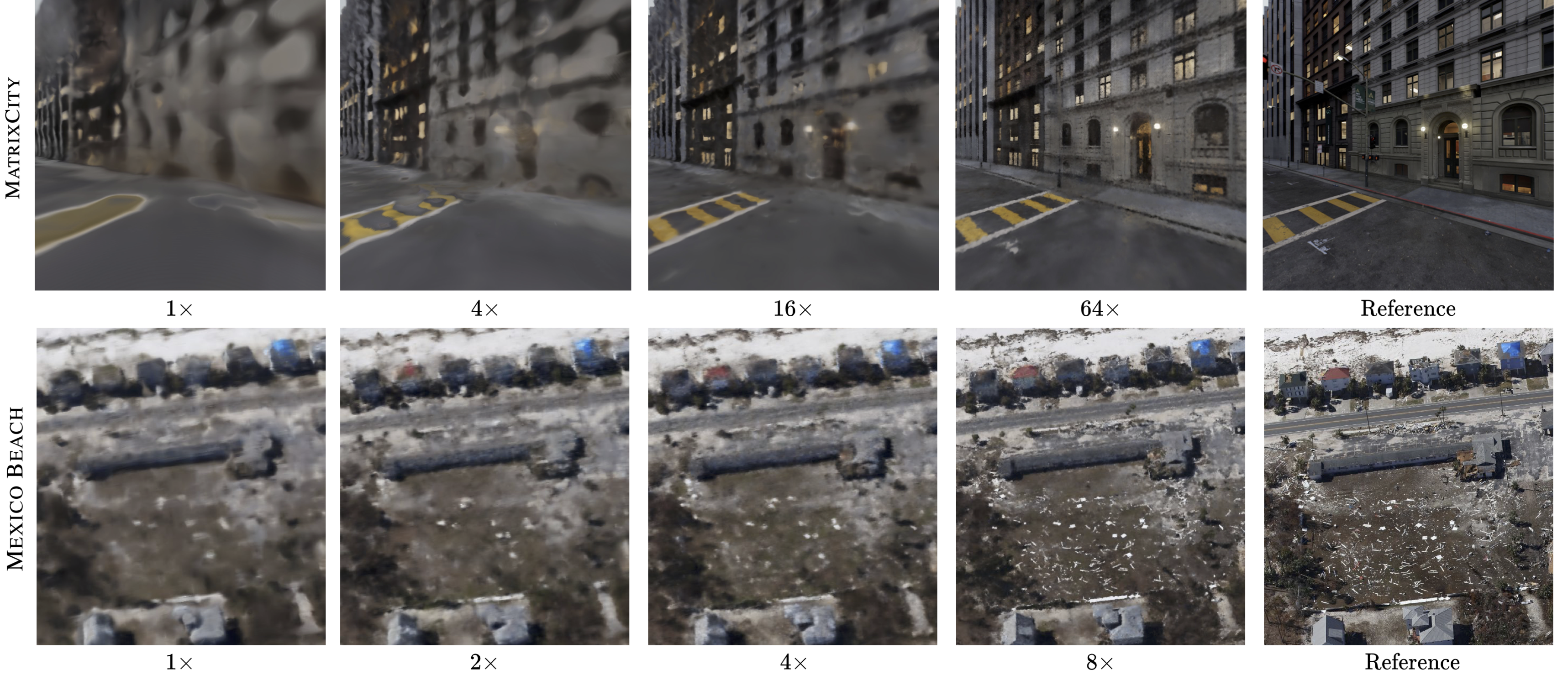}
\caption{\textbf{Scalability of Our Approach.} More GPUs allow for more learnable parameters, leading to larger model capacity with better quality.}
\label{fig:visuals_scaling_up}
\end{figure*}

\begin{figure*}[!t]
\centering
\includegraphics[width=1.00\linewidth]{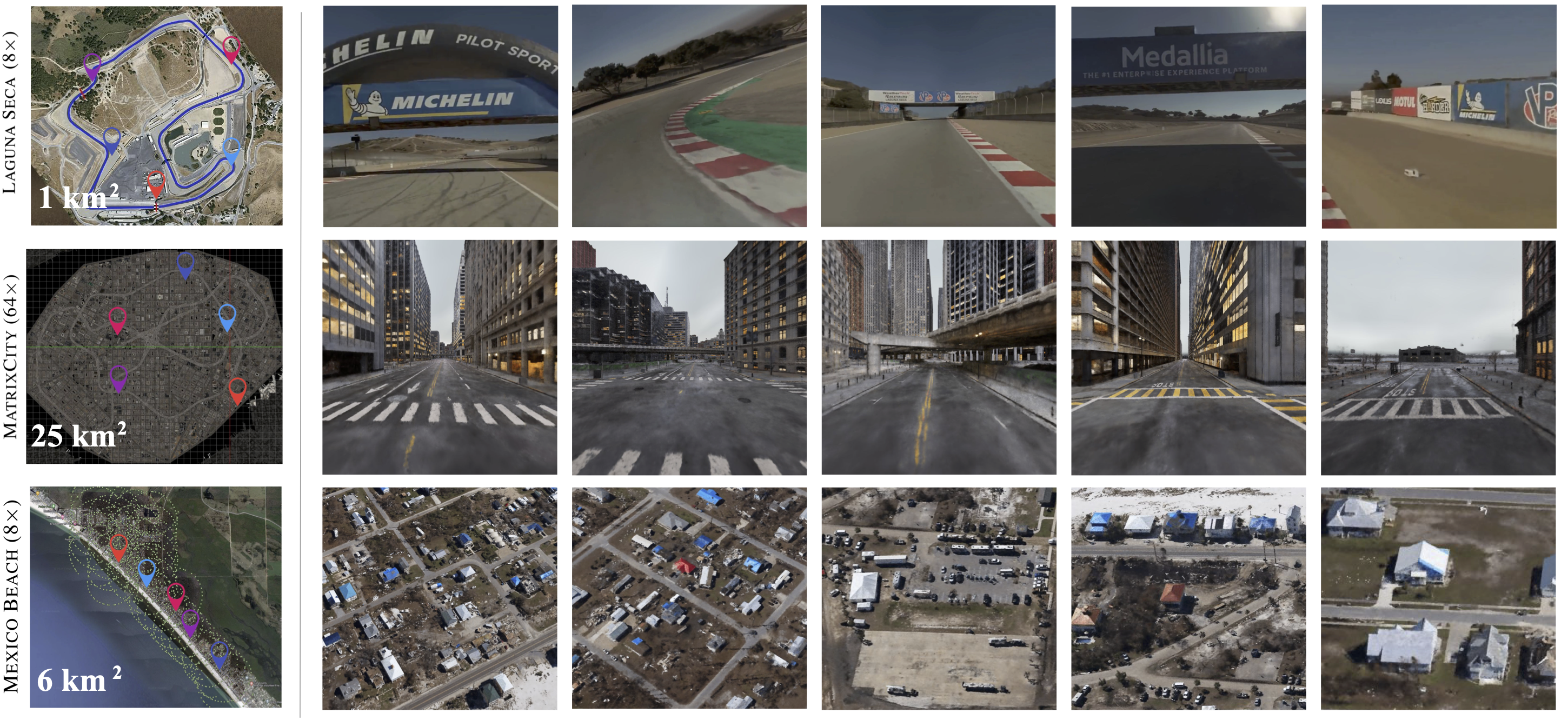}
\caption{\textbf{More Rendering Results on Large Scale Captures.} We test the robustness of our approach on larger captures with more GPUs. Please refer to the our webpage for video tours on these data.}
\label{fig:more_visuals}
\end{figure*}

We further evaluate the robustness and scalability of our approach by testing it on larger scale captures with increased GPU resources. Specifically, Figure~\ref{fig:more_visuals} showcases our novel-view rendering results on the 1km$^2$ \textsc{Laguna Seca} with 8 GPUs, the 6km$^2$ \textsc{Mexico Beach}~\cite{civilair} with 8 GPUs, and the 25km$^2$ \textsc{MatrixCity}~\cite{li2023matrixcity} with 64 GPUs. It's noteworthy that each of these captures entails billions of pixels (see Table~\ref{tab:data_stats}), posing a significant challenge to the NeRF model's capacity due to the vast amount of information being processed.

Figure~\ref{fig:visuals_scaling_up} presents qualitative results obtained using our approach, highlighting how the quality improves with the incorporation of more parameters through the utilization of additional GPUs. Please refer to our webpage for the video rendering.

\subsection{Comparison with PyTorch DDP}

\begin{figure}[!t]
\centering
\includegraphics[width=0.7\linewidth]{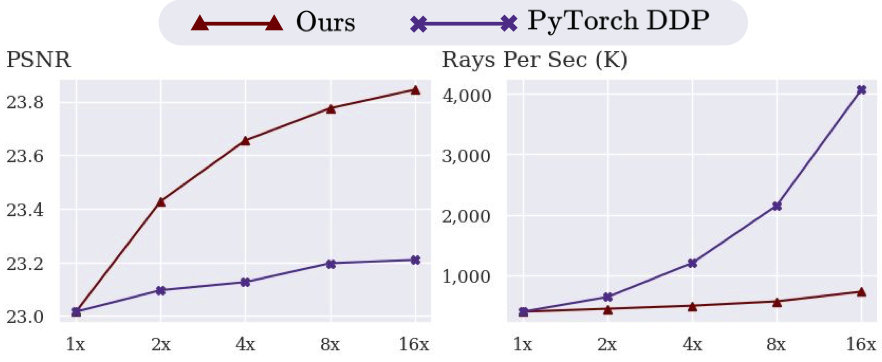}
\caption{\textbf{Comparison with PyTorch DDP on \textsc{University4}.} PyTorch Distributed Data Parallel (DDP) is designed for faster rendering by distributing rays across GPUs. In contrast, our approach distributes parameters across GPUs, scaling beyond the memory limits of single GPU in the cluster, and enabling larger model capacity for better quality.}
\label{fig:compare_ddp}
\end{figure}

Another common approach to utilize multi-GPU for NeRF is distributing rays across GPUs, \textit{e.g.}, PyTorch's Distributed Data Parallel (DDP). This method typically allows for larger batch sizes during training or faster rendering through increased parallelization. However, DDP necessitates that all GPUs host \emph{all} model parameters, thus limiting the model's capacity to the memory of a single GPU.
In contrast, our approach assigns each GPU to handle a distinct 3D tiled region, aiming to alleviate memory constraints and ensure optimal quality even for large-scale scenes. Figure~\ref{fig:compare_ddp} illustrates a comparison between our method and DDP on the \textsc{University4} dataset. In this comparison, our method employs N$\times$ more parameters while DDP trains with N$\times$ more rays on N GPUs. The substantial improvement in PSNR indicates that large-scale NeRF benefits more from increased model capacity than from training more rays, a benefit uniquely enabled by our approach.
However, DDP renders much faster than our approach due to the balanced workload created by parallelizing rays across GPUs. In contrast, our approach does not guarantee balanced workload distribution and consequently suffers from multi-GPU synchronization in run-time.

\subsection{Multi-GPU Communication}
\label{sec:abl}

\begin{figure}[!t]
\centering
\includegraphics[width=0.5\linewidth]{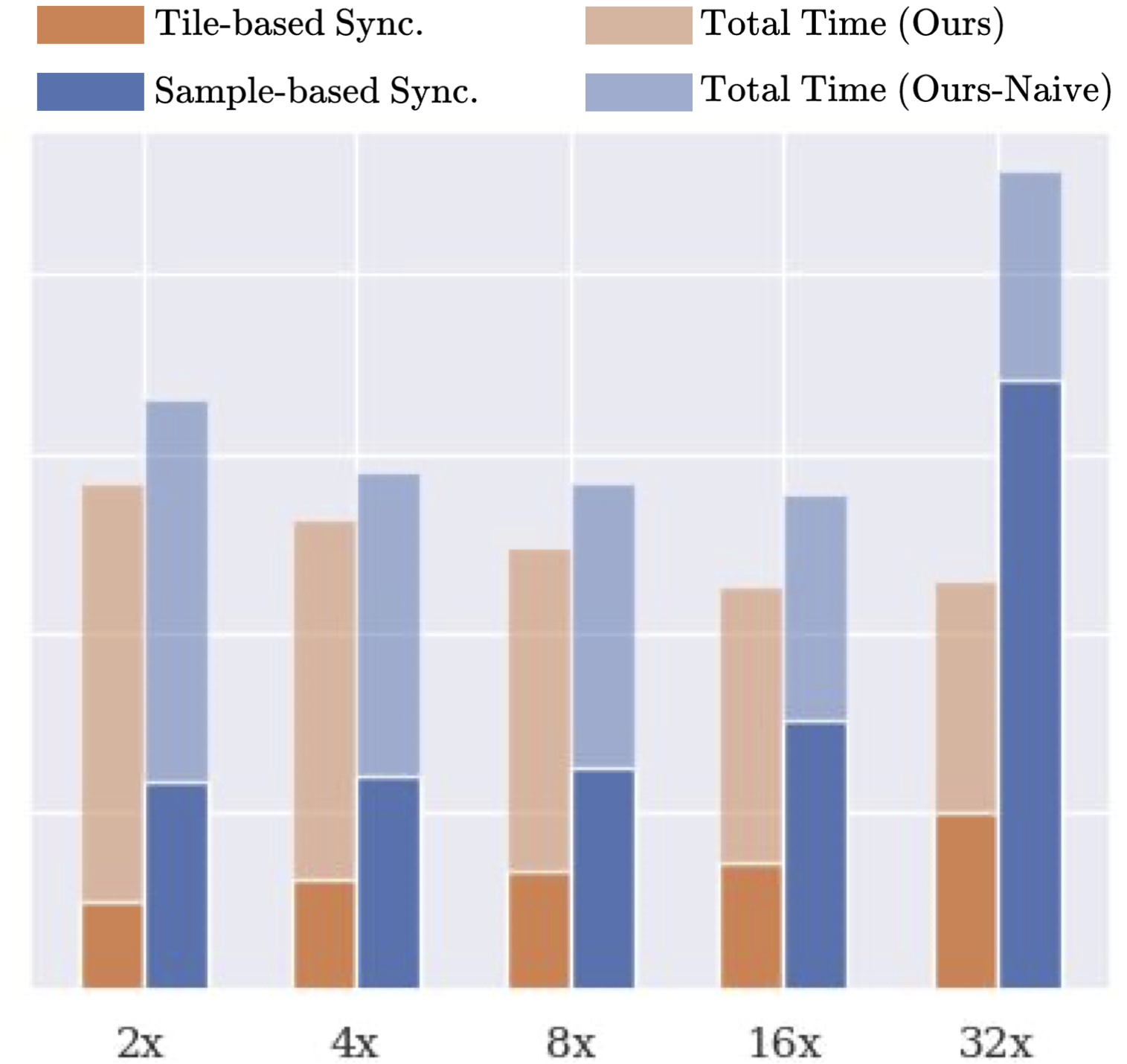}
\caption{\textbf{Synchronization Cost on \textsc{University4}}. Our partition-based volume rendering (\S~\ref{sec:method_efficient}) allows tile-based communication, which is much cheaper than the naive sample-based communication (\S~\ref{sec:method_naive}), thus enabling faster rendering.}
\label{fig:sync_time}
\end{figure}

We report the profiling results of multi-GPU communication costs on the \textsc{University4} capture in Figure~\ref{fig:sync_time}. Despite achieving a reduction in communication costs by over 2$\times$ through partition-based volume rendering (tile-based vs. sample-based synchronization), multi-GPU communication remains the primary bottleneck of our system. We attribute this to imbalanced workload distribution across GPUs, which could potentially be addressed through better spatial partitioning algorithms. We leave this optimization for future exploration.

\section{Conclusion and Limitation}
\label{sec:conclusion}
In conclusion, we revisited the existing approaches of decomposing large-scale scenes into independently trained NeRFs, and identified significant issues that impeded the effective utilization of additional computational resources (GPUs), thereby contradicting the core objective of leveraging multi-GPU setups to improve large-scale NeRF performance.
Consequently, we introduced \ShortName, a principled algorithm to efficiently harness multi-GPU setups, and enhance NeRF performance at any scale by jointly training multiple non-overlapped NeRFs. Importantly, our method does not rely on any heuristics, and adheres to scaling laws for NeRF in the multi-GPU setting across various types of data.

However, our approach still has limitations. 
Similar to any other multi-GPU distributed setup, synchronization and communication overhead is inevitable in our joint training approach, which results in a slightly slower training speed (1$\times$-1.5$\times$) compared to baselines with independent training. Additionally, while our distributed approach is agnostic to NeRF representation in theory, we have been only experimented with a popular choice, Instant-NGP~\cite{mueller2022instant}, that equips with hash grids and MLPs. It will be an interesting future work to apply the framework to other representations, even beyond the task of static scene novel-view synthesis.

% our experiments are conducted based on Instant-NGP~\cite{mueller2022instant}, a grid-based NeRF representation that places high demands on GPU memory. While in theory our joint training approach applies to other NeRF representations (\textit{e.g.}, TensoRF~\cite{chen2022tensorf}), further evidence is needed to support this claim.

% One major drawback is that joint training introduces multi-GPU synchronization in both training and rendering. Although our partition-based volume rendering alleviates this issue to some extent, it remains a significant bottleneck in the pipeline, resulting in slightly slower training speed (1$\times$-1.5$\times$) compared to baselines with independent training. Additionally, we have only demonstrated our approach with Instant-NGP~\cite{mueller2022instant}, a grid-based NeRF representation that places high demands on GPU memory. While our joint training approach theoretically applies to other NeRF representations (\textit{e.g.}, TensoRF~\cite{chen2022tensorf}), further evidence is needed to support this claim.

% communication overhead existis for anny multigpu setup, this results in a slightly slower training speed (..) but at the benefit of arifact and heuristic free nerf scaling to multiple GPUs. ours works for any nerf representations in theory. We experimented with the popular grid based nerf representations, it will be an interesting future work to apply the framework to other representations.

\section{Acknowledgement}
This project is supported in part by IARPA DOI/IBC 140D0423C0035. We would like to thank Brent Bartlett and Tim Woodard for providing and helping with processing the Mexico Beach data.


\newpage
\appendix
% \vspace{-3mm}
\section{Implementation Details}

In this section, we elaborate on the implementation intricacies of our multi-GPU NeRF representation, multi-GPU volume rendering and distortion loss, as well as the spatial partitioning strategy we employed. Additionally, we outline detailed differences between the baseline approaches and our method from the implementation perspective.

\subsection{NeRF Representation}

Our experiments are all conducted with the hash-grid based NeRF representation introduced in Instant-NGP~\cite{mueller2022instant}. In single GPU experiments, we adhere to the original configuration of Instant-NGP. This entails predicting sample density through hash encoding followed by a one-layer MLP, and predicting sample color through another two-layer MLP. The latter is conditioned on the view direction and, optionally, an appearance embedding.

In the multi-GPU joint training scenario, we assign each GPU a NeRF with its own independent hash encoding and density MLP. 
However, for the color MLP, we adopt a different approach. As the sample color is conditioned on the view direction and optionally an appearance embedding, which can be out of training distribution during novel-view rendering, independent color MLPs will lead to inconsistent color prediction given the same input. Thus we adopt parameter sharing across all GPUs for the color MLP utilizing Distributed Data Parallel (DDP), to ensure consistent interpretation of novel view direction and appearance embedding among all GPUs. Notably, when enabled during training, an appearance embedding of zero is utilized for rendering novel-view images. Similarly, when camera optimization is enabled, the camera pose embedding of each camera is also shared across GPUs.

While our experiments specifically focus on the Instant-NGP representation, we believe our approach can be generalized to many other NeRF representations. The Instant-NGP representation combines grid-based and MLP elements, suggesting that our multi-GPU distribution strategy is applicable to both grid-based and MLP-based NeRFs, as long as each NeRF is confined within a non-overlapping bounding box for proper integration. In essence, regardless of which representation is used, our approach simply increases its model capacity by deploying multiple instantiations with multi-GPU, while enabling parameter sharing across them. Consequently, all instantiations are conceptually united as a ``single NeRF'' with spatially partitioned parameters.

\subsection{Multi-GPU Volume Rendering and Distortion Loss}

% \vspace{-3mm}
\begin{algorithm}[h!]
\caption{Volume Rendering on $k$-th GPU}
\label{alg:volrend}
\KwData{
{\begin{minipage}[t]{6cm}%
 \strut
 $T_{k\veryshortarrow{k+1}}$; $C_{k\veryshortarrow{k+1}}$;
 $I=\{i_1,i_2,...i_N\}$;
 \strut
\end{minipage}%
}
}
\KwResult{$C_{1\veryshortarrow{N+1}}$;}
$T_{1\veryshortarrow{i}} \gets 1$\;
$C_{1\veryshortarrow{N+1}} \gets 0$\;
\For{$s \gets 1$ \KwTo $N$} {
    \Comment{\parbox[t]{0.65\linewidth}{\footnotesize $s$-th segment is from $i$-th GPU.}}
    $i \gets I[s]$\;
    \If{$i \neq k$}{
        \Comment{\parbox[t]{.70\linewidth}{\footnotesize Get the data from $i$-th GPU with auto-grad disabled.}}
        $T_{i\veryshortarrow{i+1}}, C_{i\veryshortarrow{i+1}} \gets \text{gather}(i)$\; 
    }
    \Comment{\parbox[t]{.70\linewidth}{\footnotesize Global composition.}}
    $C_{1\veryshortarrow{N+1}} \plusequals T_{1\veryshortarrow{i}} \times C_{i\veryshortarrow{i+1}}$\; 
    $T_{1\veryshortarrow{i}} \timesequals T_{i\veryshortarrow{i+1}}$\;
}
\end{algorithm}
% \vspace{-3mm}

% \vspace{-3mm}
\begin{algorithm}[h!]
\caption{Distortion Loss on $k$-th GPU}
\label{alg:dist}
\KwData{
{\begin{minipage}[t]{6cm}%
 \strut
 $T_{k\veryshortarrow{k+1}}$; $C_{k\veryshortarrow{k+1}}$, $I=\{i_1,i_2,...i_N\}$\;
 $A_{k\veryshortarrow{k+1}}$, $D_{k\veryshortarrow{k+1}}$; $\mathcal{L}_{k\veryshortarrow{k+1}}$;
 \strut
\end{minipage}%
}
}
\KwResult{$\mathcal{L}_{1\veryshortarrow{N+1}}$;}
$T_{1\veryshortarrow{i}} \gets 1$\;
$A_{1\veryshortarrow{i}} \gets 0$\;
$D_{1\veryshortarrow{i}} \gets 0$\;
$\mathcal{L}_{1\veryshortarrow{N+1}} \gets 0$\;
\For{$s \gets 1$ \KwTo $N$} {
    \Comment{\parbox[t]{0.65\linewidth}{\footnotesize $s$-th segment is from $i$-th GPU.}}
    $i \gets I[s]$\;
    \If{$i \neq k$}{
        \Comment{\parbox[t]{.70\linewidth}{\footnotesize Get the data from $i$-th GPU with auto-grad disabled.}}
        $T_{i\veryshortarrow{i+1}}, C_{i\veryshortarrow{i+1}} \gets \text{gather}(i)$\; 
        $A_{i\veryshortarrow{k+1}}, D_{i\veryshortarrow{i+1}}, \mathcal{L}_{i\veryshortarrow{i+1}} \gets \text{gather\_loss}(i)$\; 
    }
    \Comment{\parbox[t]{.70\linewidth}{\footnotesize Global composition.}}
    $S_{1\veryshortarrow{i}} \gets D_{k\veryshortarrow{k+1}} \times A_{1\veryshortarrow{i}} - A_{k\veryshortarrow{k+1}} \times D_{1\veryshortarrow{i}}$\;
    $\mathcal{L}_{1\veryshortarrow{N+1}} \plusequals T_{1\veryshortarrow{i}}^2 \times \mathcal{L}_{i\veryshortarrow{i+1}} + T_{1\veryshortarrow{i}} \times S_{1\veryshortarrow{i}}$\;     
    $A_{1\veryshortarrow{i}} \plusequals T_{1\veryshortarrow{i}} \times A_{i\veryshortarrow{i+1}}$\; 
    $D_{1\veryshortarrow{i}} \plusequals T_{1\veryshortarrow{i}} \times D_{i\veryshortarrow{i+1}}$\; 
    $T_{1\veryshortarrow{i}} \timesequals T_{i\veryshortarrow{i+1}}$\;
}
\end{algorithm}
% \vspace{-3mm}

Algorithm~\ref{alg:volrend} and~\ref{alg:dist} demonstrate the implementation of multi-GPU volume rendering and distortion loss on each GPU, corresponding to the formulations presented in~\S~4.3 of the main paper. Both algorithms take in locally integrated data (e.g., $T_{k\rightarrow{k+1}}$, $C_{k\rightarrow{k+1}}$) within the $k$-th GPU, and perform a global aggregation of integrated data from all GPUs. This ensures identical outputs on every GPU, but each lives in a distinct computational graph that is only differentiable with respect to it's respective NeRF parameters. During inference, data gathering is only required on a single GPU (e.g., $0$-th GPU) to render the final image, necessitating less data transfer compared to training. While we present our algorithms here as a loop over all GPUs for clarity, in practice, this is accomplished through batched asynchronous send/receive operations executed in parallel. The global composition is implemented with parallel prefix scan using the NerfAcc~\cite{li2023nerfacc} toolbox.

\subsection{Spatial Partitioning}

\begin{figure*}[!t]
\centering
\includegraphics[width=1.00\linewidth]{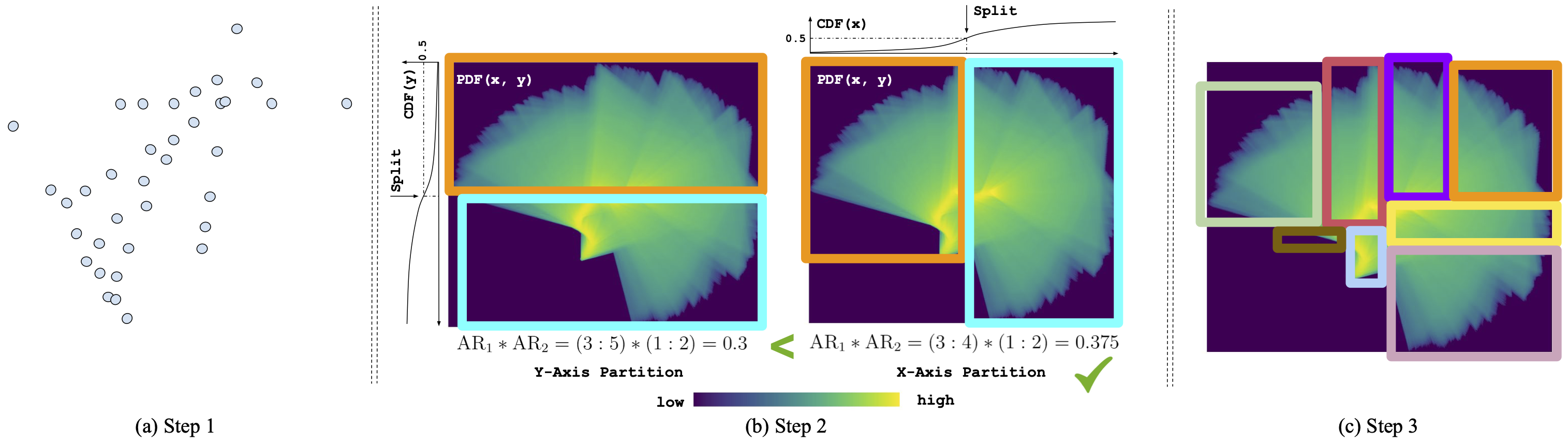}
\caption{\textbf{Our Spatial Partitioning Approach.} 
We partition the space such that each region contains similar amount of spatial content, which helps balance the compute across multi-GPUs. \textbf{(a)} Initially, we generate a point cloud either through Structure from Motion (SFM) or by discretizing training rays into samples. \textbf{(b)} Subsequently, we construct a Probability Density Function (PDF) for the point cloud and partition the space accordingly. This process is repeated for each axis, selecting the partitioning scheme that results in partitions as close to cubic as possible (denoted by $\text{AR}_i$, representing the aspect ratio for each partition). \textbf{(c)} Finally, we recursively apply step (b) for $n$ iterations to achieve $2^n$ partitions.
} 
\label{fig:partition_illustration}
\end{figure*}

\begin{figure*}[!t]
\centering
\includegraphics[width=1.00\linewidth]{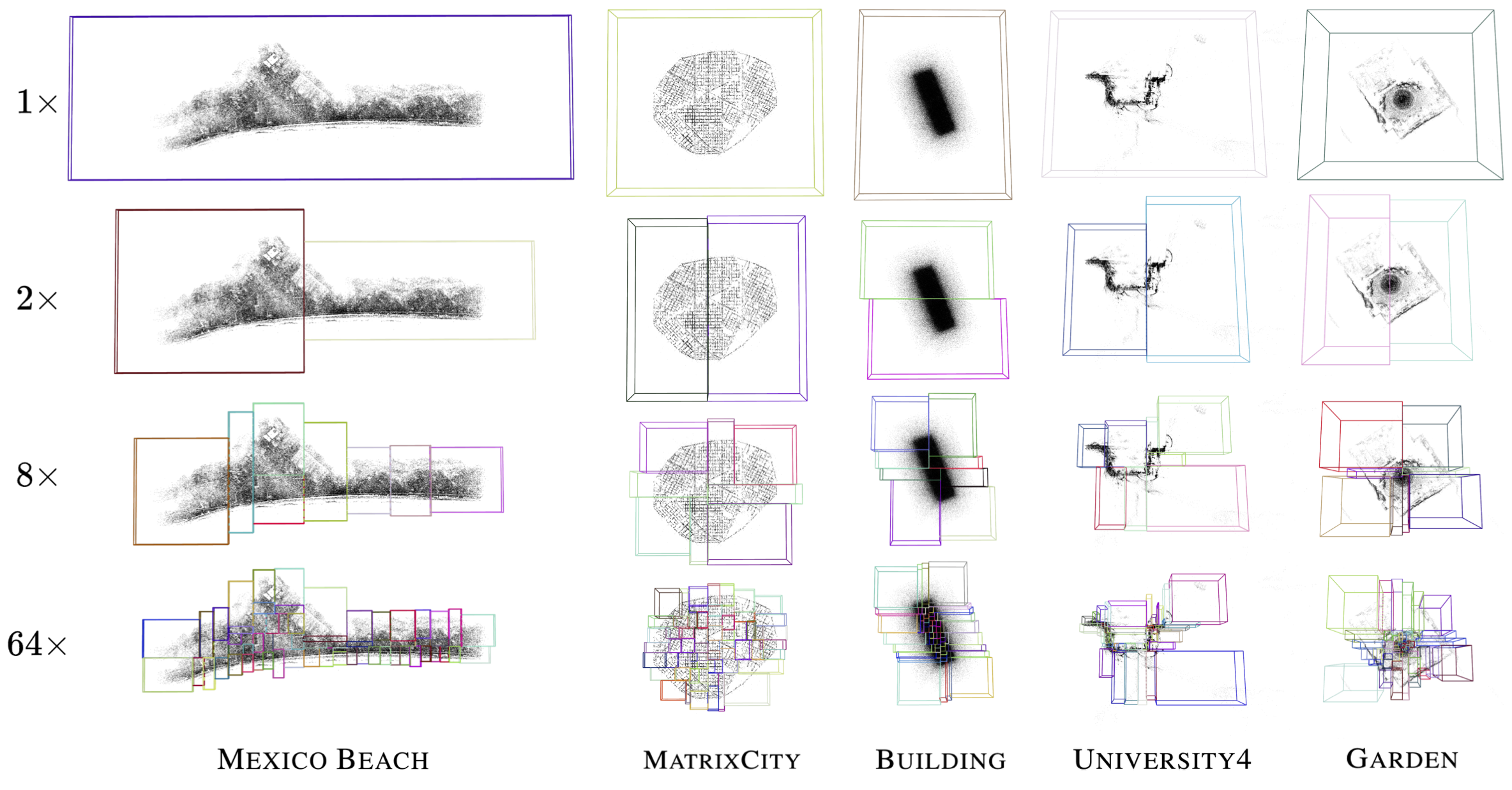}
\caption{\textbf{Our Spatial Partitioning Results.} 
Our partitioning approach is versatile and applicable to various types of captures, including drone footage (\textsc{Mexico Beach}, \textsc{Building}), street capture (\textsc{MatrixCity}, \textsc{University4}), and 360-degree object-centric capture (\textsc{Garden}). For all captures except \textsc{Building}, partitioning is based on SFM sparse points. However, due to the absence of point cloud data, the \textsc{Building} capture is partitioned based on discretizing training rays into samples.
Our partition strategy ensures an even distribution of points across tiles, thereby naturally demonstrating a level-of-detail property. This means that regions with more content or those more frequently captured will have finer bounding boxes, enhancing granularity where necessary.
} 
\label{fig:partition_results}
\end{figure*}

Figure~\ref{fig:partition_illustration} illustrate our partition scheme. Initially, we create a point cloud from either SFM or by discretizing random training rays into samples, forming a Probability Density Function (PDF) for the distribution. By computing the Cumulative Distribution Function (CDF) along each axis ${x, y, z}$ in 3D space, we identify the candidate planes where the CDF equals 0.5, signifying an optimal separation that evenly divides the space into two partitions. To create nearly cubic partitions, we choose the plane yielding partitions whose aspect ratios are as close to 1 as possible. We apply this process recursively within each partition, generating a power-of-two number of tile ($2\times, 4\times, 8\times, ...$) for our distributed NeRFs. Figure~\ref{fig:partition_results} provides a visualization on the partitions we get on various captures.

\subsection{Baselines}

Below, we detail the implementation variances between the baseline approaches (Block-NeRF~\cite{tancik2022blocknerf}, Mega-NeRF~\cite{turki2022mega}) and our method, all of which are integrated into the same system with identical configurations, including NeRF representation, spatial skipping acceleration structure, distortion loss, and multi-GPU parallel inference.

\paragraph{Ours.} Our method partitions the space into non-overlapping tiles, with each NeRF assigned to a specific tile. All NeRFs are jointly trained using all training rays.

\paragraph{Mega-NeRF.} The original Mega-NeRF paper employs uniform spatial partitioning, which may be suboptimal for free-trajectory captures~\cite{wang2023f2}. To ensure a fair comparison, we apply the same spatial partitioning scheme as our approach. However, Mega-NeRF trains each NeRF independently, necessitating background modeling for every NeRF. To address this, we employ the scene contraction method from Mip-NeRF 360~\cite{barron2022mip360}, enabling each NeRF to focus on its assigned spatial region while also modeling the background. Consistent with the original approach, we utilize only the rays that intersect with the bounding box of each tile during NeRF training.

\paragraph{Block-NeRF.} Instead of partitioning the space into tiles, the original Block-NeRF paper partitions it based on training cameras. Specifically, it divides the training data by grouping nearby cameras with some overlap, resulting in each NeRF focusing on a different spatial region during independent training. Since background modeling is required for each NeRF, and it is non-trivial to assign each NeRF a distinct bounding box from a group of cameras, scene contraction does not apply. Therefore, we utilize the same bounding box covering the entire region for all independently trained NeRFs.
\section{Math}

Here we first provide derivations of the proposed partitioned volume rendering described in~\S~4.3 in the main paper. Then we show that not only the volume rendering equation and distortion loss can be properly distributed across GPUs, there is a family of functions can be distributed in the same way.

\subsection{Derivations of Partitioned Volume Rendering}

Here we provide the full derivations of the equations in~\S~4.3, where we turn the volume rendering and distortion loss in the region of $[t_n\veryshortarrow t_f]$ into the regions of $N$ segments $[t_1\veryshortarrow t_2, t_2\veryshortarrow t_3, ..., t_N\veryshortarrow t_{N+1}]$.

\paragraph{Transmittance}

Firstly, given that
\begin{equation}
\begin{aligned}
T(t_1\veryshortarrow t) = \exp\left(-\int \limits_{t_1}^{t} \sigma(s) ds\right)
\end{aligned}
\end{equation}

We can easily see that
\begin{equation}
\begin{aligned}
\label{equ:deriv_trans_helper}
T(t_1\veryshortarrow t)
& = \exp\left(-\int \limits_{t_1}^{t} \sigma(s) ds\right) \\
& = \exp\left(-\int \limits_{t_1}^{t_k} \sigma(s) ds -\int \limits_{t_k}^{t} \sigma(s) ds\right) \\
& = \exp\left(-\int \limits_{t_1}^{t_k} \sigma(s) ds\right) \exp\left(-\int \limits_{t_k}^{t} \sigma(s) ds\right) \\
& = T(t_1\veryshortarrow t_k) T(t_k\veryshortarrow t)
\end{aligned}
\end{equation}

And similarly:
\begin{equation}
\begin{aligned}
T(t_1\veryshortarrow t_k)
& = \exp\left(-\int \limits_{t_1}^{t_k} \sigma(s) ds\right) \\
& = \exp\left(-\sum_{k=1}^{k-1}\int \limits_{t_k}^{t_{k+1}} \sigma(s) ds \right) \\
& = \prod_{k=1}^{k-1}\exp\left(-\int \limits_{t_k}^{t_{k+1}} \sigma(s) ds \right) \\
& = \prod_{k=1}^{k-1} T(t_k\veryshortarrow t_{k+1})
\end{aligned}
\end{equation}
which is the \emph{Equation 4} in the main paper.

\paragraph{Accumulated Colors, Weights and Depths} Given the volume rendering equation that accumulates colors $c(t)$ within the range of $[t_n\veryshortarrow t_f]$ along the ray:
\begin{equation}
\begin{aligned}
C(t_n\veryshortarrow t_f) = \int \limits_{t_n}^{t_f} T(t_n\veryshortarrow t) \sigma(t) c(t) dt
\end{aligned}
\end{equation}

We can derive the partitioned version of it (\emph{Equation 4} in the main paper) with the help of $T(t_1\veryshortarrow t)=T(t_1\veryshortarrow t_k) T(t_k\veryshortarrow t)$ that we just derived in Equation~\ref{equ:deriv_trans_helper}:
\begin{equation}
\begin{aligned}
C(t_1\veryshortarrow t_{N+1})
& = \int \limits_{t_1}^{t_{N+1}} T(t_1\veryshortarrow t)\sigma(t)c(t)dt \\
&= \sum_{k=1}^{N}\left[\int \limits_{t_k}^{t_{k+1}} T(t_1\veryshortarrow t)\sigma(t)c(t)dt\right] \\
&= \sum_{k=1}^{N}\left[\int \limits_{t_k}^{t_{k+1}} T(t_1\veryshortarrow t_k) T(t_k\veryshortarrow t)\sigma(t)c(t)dt\right] \\
& = \sum_{k=1}^{N}\left[T(t_1\veryshortarrow t_k)\int \limits_{t_k}^{t_{k+1}}T(t_k\veryshortarrow t)\sigma(t)c(t)dt\right] \\
& = \sum_{k=1}^{N}T(t_1\veryshortarrow t_k)C(t_k\veryshortarrow t_{k+1}) \\
\end{aligned}
\end{equation}

The accumulated weights and depths (\emph{Equation 5 and 6} in the main paper) can be derived in similar ways with the accumulated colors, thus we omit their derivations here.

\paragraph{Distortion Loss}

The original distortion loss has the form of\footnote{The distortion loss in this supplemental material has slightly different notations comparing to the equation we have in the main paper. We substitute $t_k$ with $u$ and $t_j$ with $v$ to reduce confusion in the following derivations.}
\begin{equation}
\begin{aligned}
\mathcal{L}_{dist}(t_1\veryshortarrow t_{N+1})
& = \iint \limits_{t_1}^{t_{N+1}} w(t_1\veryshortarrow u) w(t_1\veryshortarrow v) \left| u - v \right| du dv \\
\end{aligned}
\end{equation}
in which $w(t_1\veryshortarrow u) = T(t_1\veryshortarrow u)\sigma(u)$ represents the volume rendering weight for the sample at location $u$ along the ray.

First, we can break the double integral $\iint_{t_1}^{t_{N+1}}$ into two terms, $\int_{t_1}^{t_{N+1}} \int_{t_1}^{u}$ and $\int_{t_1}^{t_{N+1}} \int_{t_1}^{v}$, in which the first term covers all $u \geq v$ and the second term covers all $v \geq u$:
\begin{equation}
\begin{aligned}
\mathcal{L}_{dist}(t_1\veryshortarrow t_{N+1})
& = \iint \limits_{t_1}^{t_{N+1}} w(t_1\veryshortarrow u) w(t_1\veryshortarrow v) \left| u - v \right| du dv \\
& = \int \limits_{t_1}^{t_{N+1}} w(t_1\veryshortarrow u) \left[ \int \limits_{t_1}^{u} w(t_1\veryshortarrow v) \left( u - v \right) dv \right] du \\
& \quad + \int \limits_{t_1}^{t_{N+1}} w(t_1\veryshortarrow v) \left[ \int \limits_{t_1}^{v} w(t_1\veryshortarrow u) \left( v - u \right) du \right] dv
\end{aligned}
\end{equation}

Since $u$ and $v$ are symmetric notations, it is evident that the two terms above are equal. Thus:
\begin{equation}
\begin{aligned}
\label{equ:deriv_dist_z}
\mathcal{L}_{dist}(t_1\veryshortarrow t_{N+1}) &= 2 \int \limits_{t_1}^{t_{N+1}} w(t_1\veryshortarrow u) z(t_1\veryshortarrow u) du \\
\text{where} \quad z(t_1\veryshortarrow u) &= \int \limits_{t_1}^{u} w(t_1\veryshortarrow v) \left( u - v \right) dv
\end{aligned}
\end{equation}

Assumes the sample $u$ belongs to the $k$-th segment (i.e., $t_k \leq u \leq t_{k+1}$), then we can break the above term $z(t_1\veryshortarrow u)$ into the integral up to $t_k$ plus the integral from $t_k$ to $u$:
\begin{equation}
\begin{aligned}
\label{equ:deriv_dist_z_two_terms}
z(t_1\veryshortarrow u)
& = \int \limits_{t_1}^{u} w(t_1\veryshortarrow v) \left( u - v \right) dv \\
& = \int \limits_{t_1}^{t_k} w(t_1\veryshortarrow v) \left( u - v \right) dv + \int \limits_{t_k}^{u} w(t_1\veryshortarrow v) \left( u - v \right) dv \\
& =  z(t_1\veryshortarrow t_k) + T(t_1\veryshortarrow t_k) \cdot z(t_k\veryshortarrow u)
\end{aligned}
\end{equation}

Substituting the term $z(t_1\veryshortarrow u)$ in Equation~\ref{equ:deriv_dist_z} with Equation~\ref{equ:deriv_dist_z_two_terms}, we now have the distortion loss in two terms:
\begin{equation}
\begin{aligned}
\mathcal{L}_{dist}(t_1\veryshortarrow t_{N+1}) 
& = 2 \int \limits_{t_1}^{t_{N+1}} w(t_1\veryshortarrow u) z(t_1\veryshortarrow t_k) du \\
& \quad + 2 \cdot T(t_1\veryshortarrow t_k) \cdot \int \limits_{t_1}^{t_{N+1}} w(t_1\veryshortarrow u) z(t_k\veryshortarrow u) du
\end{aligned}
\end{equation}

Then we break the integral $\int_{t_1}^{t_{N+1}}$ along the entire ray into the summation of integral on each segment $[t_k, t_{k+1}]$:
\begin{equation}
\begin{aligned}
\label{equ:deriv:final_L}
\mathcal{L}_{dist}(t_1\veryshortarrow t_{N+1}) &= 2 \sum_{k=1}^{N} \left(\mathcal{L}_{1} + T(t_1\veryshortarrow t_k) \cdot \mathcal{L}_{2} \right) \\
\text{where} \quad \mathcal{L}_{1} &= \int \limits_{t_k}^{t_{k+1}} w(t_1\veryshortarrow u) z(t_1\veryshortarrow t_k) du \\
\mathcal{L}_{2} &= \int \limits_{t_k}^{t_{k+1}} w(t_1\veryshortarrow u) z(t_k\veryshortarrow t_u) du \\
\end{aligned}
\end{equation}

Focusing on $\mathcal{L}_{1}$, with the help of Equation~\ref{equ:deriv_dist_z} we have:
\begin{equation}
\begin{aligned}
\label{equ:deriv_L1}
\mathcal{L}_{1}
&= \int \limits_{t_k}^{t_{k+1}} w(t_1\veryshortarrow u) z(t_1\veryshortarrow t_k) du\\
&= T(t_1\veryshortarrow t_k) \cdot \int \limits_{t_k}^{t_{k+1}} w(t_1\veryshortarrow u) \left[ \int \limits_{t_k}^{t_k} w(t_1\veryshortarrow v) \left( u - v \right) dv \right] du\\
% &= \int \limits_{t_k}^{t_{k+1}} w(t_1\veryshortarrow u) \left[ \int \limits_{t_1}^{t_k} w(t_1\veryshortarrow v) dv \right] u du\\
% & \quad - \int \limits_{t_k}^{t_{k+1}} w(t_1\veryshortarrow u) \left[ \int \limits_{t_1}^{t_k} w(t_1\veryshortarrow v) v dv \right] du\\
&= T(t_1\veryshortarrow t_k) \cdot \left[ \int \limits_{t_k}^{t_{k+1}} w(t_1\veryshortarrow u) u du \right] \cdot \left[ \int \limits_{t_1}^{t_k} w(t_1\veryshortarrow v) dv \right]\\
& \quad - T(t_1\veryshortarrow t_k) \cdot \left[ \int \limits_{t_k}^{t_{k+1}} w(t_1\veryshortarrow u) du \right] \cdot \left[ \int \limits_{t_1}^{t_k} w(t_1\veryshortarrow v) v dv \right]\\
% & = D(t_k\veryshortarrow t_{k+1}) A(t_1\veryshortarrow t_k) - A(t_k\veryshortarrow t_{k+1}) D(t_1\veryshortarrow t_k)
\end{aligned}
\end{equation}

Recall that the accumulated weights and depths (\emph{Equation 5 and 6} in the main paper) have the formulations of:

\begin{equation}
\begin{aligned}
\label{equ:deriv_A}
A(t_1\veryshortarrow t_{N+1})
& = \int_{t_1}^{t_{N+1}} w(t_1\veryshortarrow u) du \\
\end{aligned}
\end{equation}

\begin{equation}
\begin{aligned}
\label{equ:deriv_D}
D(t_1\veryshortarrow t_{N+1})
& = \int_{t_1}^{t_{N+1}} w(t_1\veryshortarrow u) u du \\
\end{aligned}
\end{equation}

Substituting the terms in Equation~\ref{equ:deriv_L1} with the above two equations we get the \emph{Equation 9} in the main paper:
\begin{equation}
\begin{aligned}
\label{equ:deriv:final_L1}
\mathcal{L}_{1} &= T(t_1\veryshortarrow t_k) S(t_1\veryshortarrow t_k)
\end{aligned}
\end{equation}
in which the $S(t_1\veryshortarrow t_k)$ is defined as:
\begin{equation}
\label{equ:s}
\begin{aligned}
S(t_1\veryshortarrow t_k)
 = D(t_k\veryshortarrow t_{k+1})A(t_1\veryshortarrow t_k) 
 - A(t_k\veryshortarrow t_{k+1})D(t_1\veryshortarrow t_k)
\end{aligned}
\end{equation}

Now focusing on $\mathcal{L}_{2}$, we can see that:
\begin{equation}
\begin{aligned}
\label{equ:deriv:final_L2}
\mathcal{L}_{2} 
&= \int \limits_{t_k}^{t_{k+1}} w(t_1\veryshortarrow u) z(t_k\veryshortarrow t_u) du \\
&= T(t_1\veryshortarrow t_k) \int \limits_{t_k}^{t_{k+1}} w(t_k\veryshortarrow u) z(t_k\veryshortarrow t_u) du \\
&= T(t_1\veryshortarrow t_k) \cdot \frac{\mathcal{L}_{dist}(t_k\veryshortarrow t_{k+1})}{2}
\end{aligned}
\end{equation}

Putting together the above Equations~\ref{equ:deriv:final_L},~\ref{equ:deriv:final_L1},~\ref{equ:deriv:final_L2}, we then have derived the \emph{Equation 7} in the main paper:
\begin{equation}
\begin{aligned}
\mathcal{L}_{dist}(t_1\veryshortarrow t_{N+1}) 
% &= 2 \sum_{k=1}^{N} \left(\mathcal{L}_{1} + T(t_1\veryshortarrow t_k) \cdot \mathcal{L}_{2} \right) \\
 = 2\sum_{k=1}^{N}T(t_1\veryshortarrow t_k)S(t_1\veryshortarrow t_k) 
 + \sum_{k=1}^{N}T(t_1\veryshortarrow t_k)^2 \mathcal{L}_{dist}(t_k\veryshortarrow t_{k+1})
\end{aligned}
\end{equation}

\subsection{General Derivations for More NeRF-related Functions}

In this section we prove that there is a family of integral functions defined in the range of $[t_1, t_{N+1}]$ can be rewritten into the form of sum product on the integrals of each individual segment $[t_k, t_{k+1}]$, which makes them suitable to be distributed across multiple GPUs using our approach  (\ie first calculate each segment independently within each GPU, then accumulate only per-ray data across GPUs).

\begin{equation}
\begin{aligned}
\mathcal{F}(t_1 \veryshortarrow t_{N+1}) = \sum_{i=1}^{N}(\prod_{j=1}^{N} H_{ij}(t_j\veryshortarrow t_{j+1}))
\end{aligned}
\end{equation}
which we will call them \emph{breakable} integrals.

A simple example of a breakable integral is:
\begin{equation}
\begin{aligned}
\mathcal{F}(t_1 \veryshortarrow t_{N+1}) 
= \int_{t_1}^{t_{N+1}} f(t) dt 
= \sum_{i=1}^{N} \mathcal{F}(t_i \veryshortarrow t_{i+1})
\end{aligned}
\end{equation}
which corresponds to $H_{ij} = \begin{bmatrix}
    \mathcal{F} & & \\
    & \ddots & \\
    & & \mathcal{F}
  \end{bmatrix}$.

% A breakable integral has some nice properties:
% \begin{equation}
% \begin{aligned}
% \mathcal{F}(t_1 \veryshortarrow t_{N+1}) 
% & = \sum_{i=1}^{N}(\prod_{j=1}^{N} H_{ij}(t_j\veryshortarrow t_{j+1})) \\
% & = \sum_{i=1}^{N - 1}(\prod_{j=1}^{N-1} H_{ij}(t_j\veryshortarrow t_{j+1}) \cdot H_{iN}(t_j\veryshortarrow t_{j+1})) + \prod_{j=1}^{N} H_{Nj}(t_j\veryshortarrow t_{j+1})
% \end{aligned}
% \end{equation}

\subsubsection{Properties and Proofs}

A breakable integral has several nice properties:

\paragraph{Property 1} \emph{A breakable integral multiplies, adds or subtracts a breakable integral is still a breakable integral.} 

% \paragraph{Property 2} \emph{If $\mathcal{F}(t_1 \veryshortarrow t_{N+1})$ is a breakable integral, then applying any non-integral operation $U$ to it remains a breakable integral: $U(\mathcal{F}(t_1 \veryshortarrow t_{N+1}))$} 

% \noindent\textbf{Proof:} skip.

\paragraph{Property 2} \emph{If $\mathcal{F}(t_1 \veryshortarrow t_{N+1})$ is a breakable integral, then $\mathcal{A}(t_1 \veryshortarrow t_{N+1})$ is also a breakable integral when:}
\begin{equation}
\begin{aligned}
\mathcal{A}(t_1 \veryshortarrow t_{N+1}) 
& = \int_{t_1}^{t_{N+1}} \mathcal{F}(t_1 \veryshortarrow t)f(t)dt \\
\end{aligned}
\end{equation}

\noindent{Proof:} 

\
\begin{equation}
\begin{aligned}
\mathcal{A}(t_1 \veryshortarrow t_{N+1}) 
 = \int_{t_1}^{t_{N+1}} \mathcal{F}(t_1 \veryshortarrow t)f(t)dt 
 = \sum_{k=1}^{N} \int_{t_k}^{t_{k+1}} \mathcal{F}(t_1 \veryshortarrow t)f(t)dt \\
\end{aligned}
\end{equation}
In which we have
\begin{equation}
\begin{aligned}
& \mathcal{F}(t_1 \veryshortarrow t)  \\
& = \sum_{i=1}^{k-1}(\prod_{j=1}^{k-1} H_{ij} (t_j\veryshortarrow t_{j+1}) \cdot H_{ik}(t_k\veryshortarrow t)) + \prod_{j=1}^{k-1} H_{kj}(t_j\veryshortarrow t_{j+1}) \cdot H_{kj}(t_k\veryshortarrow t) \\
& = S_1 + S_2
\end{aligned}
\end{equation}

Focusing on $S_1$, we have
\begin{equation}
\begin{aligned}
\mathcal{A}_1(t_1 \veryshortarrow t_{N+1}) 
& = \sum_{k=1}^{N} \int_{t_k}^{t_{k+1}} S_1 f(t)dt \\
& = \sum_{k=1}^{N} \int_{t_k}^{t_{k+1}} \left[ \sum_{i=1}^{k-1}(\prod_{j=1}^{k-1} H_{ij}(t_j\veryshortarrow t_{j+1}) \cdot H_{ik}(t_k\veryshortarrow t)) \right] f(t)dt \\
& = \sum_{k=1}^{N} \sum_{i=1}^{k-1}(\prod_{j=1}^{k-1} H_{ij}(t_j\veryshortarrow t_{j+1}) \cdot \left[ \int_{t_k}^{t_{k+1}} H_{ik}(t_k\veryshortarrow t) f(t)dt \right])   \\
& = \sum_{k=1}^{N} \sum_{i=1}^{k-1}(\prod_{j=1}^{k-1} H_{ij}(t_j\veryshortarrow t_{j+1}) \cdot \hat{H}_{ik}(t_k\veryshortarrow t_{k+1}))   \\
\end{aligned}
\end{equation}

Focusing on $S_2$, we have
\begin{equation}
\begin{aligned}
\mathcal{A}_2(t_1 \veryshortarrow t_{N+1}) 
& = \sum_{k=1}^{N} \int_{t_k}^{t_{k+1}} S_2 f(t)dt \\
& = \sum_{k=1}^{N} \int_{t_k}^{t_{k+1}} \left[ \prod_{j=1}^{k-1} H_{kj}(t_j\veryshortarrow t_{j+1}) \cdot H_{kj}(t_k\veryshortarrow t) \right] f(t)dt \\
& = \sum_{k=1}^{N} \prod_{j=1}^{k-1} H_{kj}(t_j\veryshortarrow t_{j+1}) \cdot \left[ \int_{t_k}^{t_{k+1}} H_{kj}(t_k\veryshortarrow t) f(t)dt \right] \\
& = \sum_{k=1}^{N} \prod_{j=1}^{k-1} H_{kj}(t_j\veryshortarrow t_{j+1}) \cdot \hat{H}_{kj}(t_k\veryshortarrow t_{k+1}) \\
\end{aligned}
\end{equation}
Thus $\mathcal{A}(t_1 \veryshortarrow t_{N+1}) = \mathcal{A}_1(t_1 \veryshortarrow t_{N+1}) + \mathcal{A}_2(t_1 \veryshortarrow t_{N+1})$ also belongs to the family of breakable integrals.

\subsubsection{Examples}

Here we show both the volume rendering equation and distortion loss can be trivially proved as breakable integrals using the above properties.

\paragraph{Transmittance} The \emph{transmittance} belongs to this family because:
\begin{equation}
\begin{aligned}
T(t_1 \veryshortarrow t_{N+1}) 
 = \exp\left(-\int_{t_1}^{t_{N+1}}\sigma(t)dt\right)
 = \prod_{i=1}^{N} T(t_i \veryshortarrow t_{i+1})
\end{aligned}
\end{equation}

\paragraph{Volume Rendering} The volume rendering has the formulation of:

\begin{equation}
\begin{aligned}
C(t_1 \veryshortarrow t_{N+1}) 
& = \int_{t_1}^{t_{N+1}} T(t_1 \veryshortarrow t) \sigma(t) c(t) dt \\
\end{aligned}
\end{equation}

Given that the transmittance $T(t_1 \veryshortarrow t)$ belongs to this family, and the property 2 of this family, we know that volume rendering equation also belongs to this family.

\paragraph{Distortion Loss} The distortion loss has the formulation of:
\begin{equation}
\begin{aligned}
\mathcal{L}_{dist}(t_1\veryshortarrow t_{N+1})
& = \iint \limits_{t_1}^{t_{N+1}} w(t_1\veryshortarrow u) w(t_1\veryshortarrow v) \left| u - v \right| du dv \\
& = 2 \int \limits_{t_1}^{t_{N+1}} w(t_1\veryshortarrow u) \left[ \int \limits_{t_1}^{u} w(t_1\veryshortarrow v) \left( u - v \right) dv \right] du \\
& = 2 \int \limits_{t_1}^{t_{N+1}} T(t_1\veryshortarrow u) \sigma(u) \left[ \int \limits_{t_1}^{u} T(t_1\veryshortarrow v) \sigma(v) \left( u - v \right) dv \right] du \\
& = 2 \int \limits_{t_1}^{t_{N+1}} T(t_1\veryshortarrow u) \sigma(u) S(t_1 \veryshortarrow u) du \\
\end{aligned}
\end{equation}

Similar to the volume rendering equation, from property 2 we know that the formula within the bracket belongs to this family:
\begin{equation}
\begin{aligned}
S(t_1 \veryshortarrow u) = \int \limits_{t_1}^{u} T(t_1\veryshortarrow v) \sigma(v) \left( u - v \right) dv
\end{aligned}
\end{equation}

From property 1 we know $T(t_1 \veryshortarrow u) S(t_1 \veryshortarrow u)$ also belongs to this family. Then apply property 2 again we can see that $\mathcal{L}_{dist}(t_1\veryshortarrow t_{N+1})$ also belongs to this family.

\bibliographystyle{splncs04}
\bibliography{main}
\end{document}